\pdfoutput=1

\documentclass[11pt]{article}

\usepackage[final]{acl}

\usepackage{times}
\usepackage{latexsym}

\usepackage[T1]{fontenc}

\usepackage[utf8]{inputenc}

\usepackage{microtype}

\usepackage{inconsolata}

\usepackage{graphicx}

\usepackage{hhline}
\usepackage{amsmath}
\usepackage{amssymb}
\usepackage{multirow}
\usepackage{arydshln}
\usepackage{enumitem}
\usepackage{multicol}
\usepackage{xspace}
\usepackage{wasysym}
\usepackage{amssymb}
\usepackage{pifont}
\usepackage{indentfirst}
\usepackage{wrapfig}
\usepackage{subcaption}
\usepackage{booktabs}

\usepackage[most]{tcolorbox}

\newcommand{\cmark}{\ding{51}}%
\newcommand{\xmark}{\ding{55}}%

\newcommand{\autorefp}[1]{(\autoref{#1})}

\newcommand{\autopromptref}[1]{\hyperref[#1]{Prompt~\ref*{#1}}}

\newcommand{\logo}[0]{\raisebox{-.2\height}{\includegraphics[width=.04\textwidth]{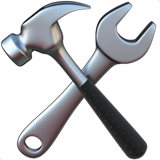}}} 

\newcommand{\dbq}[0]{\textnormal{"}\xspace}
\def\name{TAPS\xspace}
\def\extag{\textsc{Ext-Tag}}
\def\jointtag{\textsc{Joint-tag}}

\def\codellamasin{\texttt{CodeLlama-Inst}\xspace}
\def\codellamas{\texttt{CodeLlama}\xspace}

\def\mistralthrees{\texttt{Mistral-3}\xspace}
\def\mistralthreesin{\texttt{Mistral-3-Inst}\xspace}
\def\llamatwos{\texttt{Llama-2}\xspace}
\def\llamatwoschat{\texttt{Llama-2-Chat}\xspace}
\def\llamathreeschat{\texttt{Llama-3-Inst}\xspace}
\def\llamathrees{\texttt{Llama-3}\xspace}
\def\gptb{\texttt{GPT4o}\xspace}

\def\olmos{\texttt{OLMo-2-7B-Inst}\xspace}

\title{\logo  \hspace{0.3em}TAPS: Tool-Augmented Personalisation via Structured Tagging}

\author{Ekaterina Taktasheva \textnormal{and} Jeff Dalton \\
   University of Edinburgh \\
  \texttt{\{e.taktasheva, jeff.dalton\}@ed.ac.uk} \\}

\begin{document}
\maketitle
\begin{abstract}
Recent advancements in tool-augmented large language models have enabled them to interact with external tools, enhancing their ability to perform complex user tasks. However, existing approaches overlook the role of personalisation in guiding tool use. This work investigates how user preferences can be effectively integrated into goal-oriented dialogue agents. Through extensive analysis, we identify key weaknesses in the ability of LLMs to personalise tool use. To this end, we introduce \name, a novel solution that enhances personalised tool use by leveraging a structured tagging tool and an uncertainty-based tool detector. \name significantly improves the ability of LLMs to incorporate user preferences, achieving the new state-of-the-art for open source models on the NLSI task\footnote{The code is available at \href{https://github.com/grill-lab/taps}{github.com/grill-lab/taps}.}.
\end{abstract}

\section{Introduction}

Successfully completing complex user tasks through conversation remains a fundamental challenge for goal-oriented dialogue agents. Consider a user interacting with a task assistant to book a last-minute flight. To effectively assist the user, the system must (i) retrieve real-time flight availability, (ii) find the flight that fits user constraints, including airline, layover, and time preferences, (iii) and execute the booking seamlessly, possibly across multiple platforms. Despite their success in many areas, Large Language Models (LLMs) are still unable to fulfil these requirements on their own, and there have been many attempts to address these challenges throughout the years (\citealt[inter alia]{goel2018flexible, muise2019planning, agarwal2022building}).

Recently, a growing number of studies have emerged on tool-augmented language models (TALMs), allowing LLMs to access real-world APIs to perform a wide range of tasks \citep{parisi2022talmtoolaugmentedlanguage, schick2023toolformerlanguagemodelsteach}. Tool use has enabled the development of autonomous goal-oriented agents capable of interacting with real-world environments and accessing external data to seamlessly plan and execute complex user tasks \citep{mialon2023gaiabenchmarkgeneralai, qin2023toolllmfacilitatinglargelanguage, liu2024agentbench}. Although there have been efforts to incorporate tool use into conversational agents \citep{farn2023tooltalkevaluatingtoolusageconversational, li-etal-2023-api, lu2024toolsandboxstatefulconversationalinteractive}, most of the research in the area neglects conversational history and user preferences. Recognising these can enhance the user experience by tailoring the responses to individual users and improving the relevance and efficiency of task execution, especially in complex and dynamic environments. \citet{moghe-etal-2024-interpreting} attempt to bridge this gap by introducing the Natural Language Standing Instructions dataset (NLSI). To the best of our knowledge, it is the first work that addresses the problem of personalisation in TALMs, enabling more coherent and context-aware tool use through  \textit{standing instructions}, phrases that prescribe model behaviour based on the specific scenario. While the work provides a strong basis for further research on tool use personalisation, it focuses on dataset construction and provides only simple baselines.

In this work, we ask \textit{how we can effectively leverage user preferences to personalise and enhance user-agent interactions}. We conduct an extensive behavioural analysis of commonly used LLMs on the NLSI dataset and demonstrate their limited ability to accurately infer tool calls in the presence of user preferences, leading to semantic errors, missing arguments, and hallucinations. We hypothesise that introducing a high-quality intermediate representation between natural language and code can significantly enhance model performance and minimise said errors. To this end, we propose \textbf{\name} -- \textbf{T}ool-\textbf{A}ugmented  \textbf{P}ersonalisation via  \textbf{S}tructured Tagging, the first solution that leverages a structured tagging tool for data augmentation as well as an internal tool detection mechanism for personalised tool use in a dialogue setting.

Our \textit{contributions} are: \textbf{(i)} we analyse common LLMs' performance on the personalised tool-use task and identify their current weaknesses; \textbf{(ii)} we propose  structured tagging, an annotation scheme that bridges natural language and API calls by hierarchically marking functions and their arguments within user preferences, \textbf{(iii)} we introduce \name, a tuning-free approach that uses a structured tagging tool and an uncertainty-based tool detector to facilitate integration of user preferences into tool-augmented goal-oriented dialogue agents;
\textbf{(iv)} we demonstrate that our method improves the effectiveness of LLMs on the task, achieving state-of-the-art results for open-source models on the interpretation subtask of NLSI with an increase of +16.5\% in exact match (EM) and +16.9\% in F1. Our findings suggest \name's potential for generalisation to other goal-oriented tasks, where reductions in errors such as hallucinations and missing arguments could improve system reliability and user experience. With this work, we hope to inspire future research on tool-use personalisation.

\section{Task Setup}
\subsection{Task Definition}

\begin{figure}[t!]
    \centering
    \includegraphics[width=\linewidth]{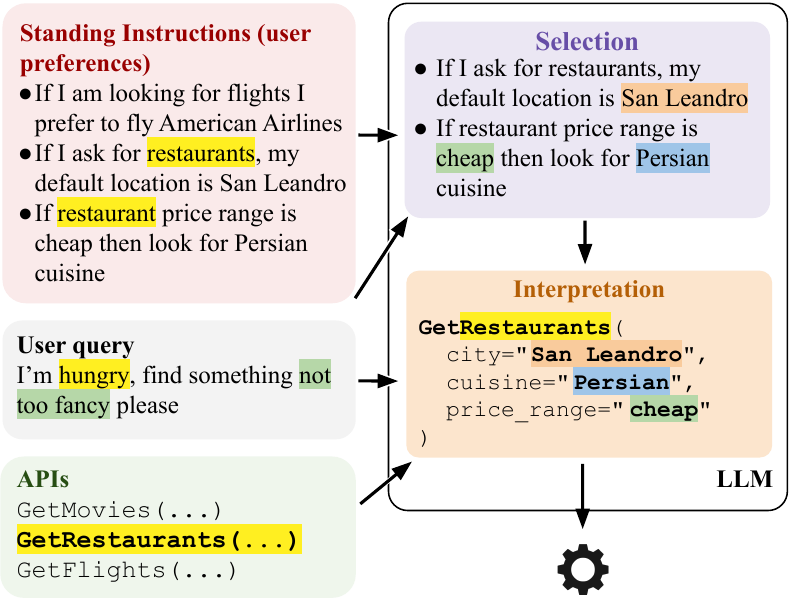}
    \caption{Example of the NLSI task. Given a user query and user-specific list of preferences, and API documentation, the model has to parse the input into structured output. The model has to (i) select, which preferences are relevant for the current query and (ii) interpret the utterance into one or several API calls. The diagram is a replica of Figure 1 from \citet{moghe-etal-2024-interpreting}.}
    \label{fig:nlsi_pipeline}
\end{figure}

The NLSI task is defined as follows. Given a user query, standing instructions, and API documentation, an agent must generate up to three API calls to fulfil the user request  \autorefp{fig:nlsi_pipeline}. The standing instructions constitute the user profile -- their preferences regarding different aspects, e.g., favourite cuisine, preferred airline, or music taste. The task requires complex reasoning to integrate query details with user preferences to generate appropriate API calls. Ultimately, the task consists of two subtasks: \textit{selection}, identifying the subset of instructions relevant to the current query; and \textit{interpretation}, generation of API calls to perform the user task using the user query, user profile, and API documentation.

This work focuses on the interpretation subtask, which is crucial for improving LLMs' ability to handle contextualised tool use -- a key challenge in real-world applications. Successful interpretation requires an agent to understand the user intent, reason over the conversation and user profile, and identify the appropriate APIs, necessary arguments, and their values. To ensure a controlled evaluation, we provide LLMs with the correct selected standing instructions, allowing them to access only the relevant user profile information.

\subsection{Evaluation}

We follow the evaluation setup, described in \citet{moghe-etal-2024-interpreting} to assess model performance. We convert each API call into (function name, argument name, value) triplets, or slots, to compute the metrics and report exact match (EM), slot-wise F1, precision, and recall. 

\subsection{Behaviour Analysis} \label{sec:behaviour}

\begin{table}[ht!]
    \centering
    \resizebox{\linewidth}{!}{
    \begin{tabular}{llrccc}
    \toprule
    \textbf{Model} & \textbf{Source} & \textbf{Size}  & \textbf{Instr.-Tuned} & \textbf{Tools}\\
    \midrule

        \href{https://huggingface.co/codellama/CodeLlama-7b-hf}{\codellamas}  & \multirow{2}{*}{\citet{rozière2024codellamaopenfoundation}} & 7B & \xmark & \xmark \\
         
       \href{https://huggingface.co/codellama/CodeLlama-7b-Instruct-hf}{\codellamasin}  &  & 7B & \cmark & \xmark \\

       \vspace{-1em}\\
        \hdashline
        \vspace{-1em}\\

        \href{https://huggingface.co/meta-llama/Llama-2-7b-hf}{\llamatwos} & \multirow{2}{*}{\citet{touvron2023llama2openfoundation}} & 7B & \xmark & \xmark \\

        \href{https://huggingface.co/meta-llama/Llama-2-7b-Chat-hf}{\llamatwoschat} & & 7B & \cmark & \xmark \\

        \vspace{-1em}\\
        \hdashline
        \vspace{-1em}\\

        \href{https://huggingface.co/meta-llama/Meta-Llama-3-8B}{\llamathrees} & \multirow{2}{*}{\citet{dubey2024llama3herdmodels}} & 8B & \xmark & \xmark \\
        
       \href{https://huggingface.co/meta-llama/Meta-Llama-3-8B-Instruct}{\llamathreeschat} &  & 8B & \cmark & \xmark\\

       \vspace{-1em}\\
        \hdashline
        \vspace{-1em}\\

        \href{https://huggingface.co/mistralai/Mistral-7B-v0.3}{\mistralthrees} & \multirow{2}{*}{\citet{jiang2023mistral7b}}  & 7B & \xmark & \cmark\\
        
       \href{https://huggingface.co/mistralai/Mistral-7B-Instruct-v0.3}{\mistralthreesin} &   & 7B & \cmark & \cmark\\
       
       \vspace{-1em}\\
        \hdashline
        \vspace{-1em}\\

        \href{https://huggingface.co/allenai/OLMo-2-1124-7B-Instruct}{\olmos} & \citet{olmo20242olmo2furious} & 7B & \cmark & \xmark\\

       \vspace{-1em}\\
        \hdashline
        \vspace{-1em}\\
       \href{https://platform.openai.com/docs/models/gpt-4-turbo-and-gpt-4}{\gptb} & \citet{openai2024gpt4technicalreport} & unk & \cmark & \cmark\\
        \bottomrule
    \end{tabular}}
    \caption{LLMs used in our work.}
    \label{tab:baseline_models}
\end{table}

The challenge of NLSI is incorporating several aspects: models must not only accurately identify the users' intended task but also select relevant information from both the current user query and the user profile, and effectively utilise it to generate the appropriate API call. An additional complexity arises from the limited availability of training data, which significantly constrains our ability to use learnable methods to solve this task.

\citet{moghe-etal-2024-interpreting} evaluate various language models on NLSI but focus on a simple in-context learning (ICL) setting. We extend this analysis by investigating the behaviour of common LMs, summarised in \autoref{tab:baseline_models}. Our experiments prioritise 7B/8B models to balance efficiency in low-resource settings and latency -- critical factors for interactive task assistants -- while recognising that larger models do not universally yield proportional performance gains despite their significantly higher resource demands. We compare our approach to \gptb (\texttt{gpt-4o-2024-08-06}), a significantly larger model, to assess capability and computational cost trade-offs. We follow \citeauthor{moghe-etal-2024-interpreting}'s evaluation setup, using their prompt in 1-shot setting (see \autoref{app:prompts}) and report results in \autoref{tab:baseline_perf}.

\begin{table}[t!]
    \centering
    \small
    \begin{tabular}{lcccc}
        \toprule
        \textbf{Model} & \textbf{EM} & \textbf{F1} & \textbf{Prec.} & \textbf{Rec.} \\
        \midrule
        \codellamas & 16.3 & 55.8 & 66.9 & 49.5 \\
        \codellamasin & 18.1 & 57.0 & 68.3 & 49.7 \\ 
        \llamatwos & 10.3 & 51.0 & 51.3 & 52.0 \\
        \llamatwoschat & 10.3 & 45.6 & 53.2 & 41.7 \\
        \llamathrees & 10.1 & 52.2 & 47.5 & 69.3 \\
        \llamathreeschat & \underline{32.5} & \underline{70.3} & \underline{68.5} & \underline{77.97} \\
        \mistralthrees & 9.7 & 54.4 & 50.1 & 66.7 \\
        \mistralthreesin & 32.7 & 65.5 & 67.6 & 65.5 \\
        \olmos & 10.8 & 43.0 & 44.6 & 46.4 \\
        \gptb & \textbf{50.4} & \textbf{84.4} & \textbf{84.4} & \textbf{87.2}\\

         \bottomrule
    \end{tabular}
    \caption{Comparison of baseline models on the NLSI test set. \textbf{EM}: exact match. \textbf{F1}: Slot-wise F1 score. \textbf{Prec.}: precision. \textbf{Rec.}: recall. All scores are in \%. Best performance is in \textbf{bold}, second best is \underline{underlined}.}
    \label{tab:baseline_perf}
\end{table}

\subsubsection{Model Comparison}

\paragraph{Effect of Model Size} 
\gptb demonstrates the highest scores across all evaluated metrics, suggesting some innate ability to infer API calls from user queries given their preferences. 
All smaller open-source models underperform significantly, highlighting the need for better and more effective interpretation techniques. 

\paragraph{Pre-Training and Post-Training Effects} 
A comparison of instruction-tuned models with their base counterparts shows that instruction fine-tuning can offer modest performance gains. However, the inferior performance of the instruction-optimised \llamatwoschat relative to its base version indicates that instruction fine-tuning does not universally result in improvements and may sometimes impede performance. 
Notably, we did not optimise the prompts for each model, which could affect model performance and lead to sub-optimal results. The significant drop in the scores of \texttt{CodeLlama} and \texttt{Llama-2} models compared to others implies that optimising LLMs for tool use enhances their ability to handle complex interpretation tasks, allowing them to better integrate various input sources and produce accurate function calls.

The substantial gap between the EM and F1 scores across all models shows that while they can produce plausible API calls, they struggle to accurately incorporate all necessary data when translating natural language into executable code. Given the lower scores of some models, we focus on \mistralthreesin, \llamathreeschat, and \gptb in our further experiments.

\subsubsection{Effect of Example Complexity}

\begin{figure}[h]
        \centering
        \includegraphics[width=\linewidth]{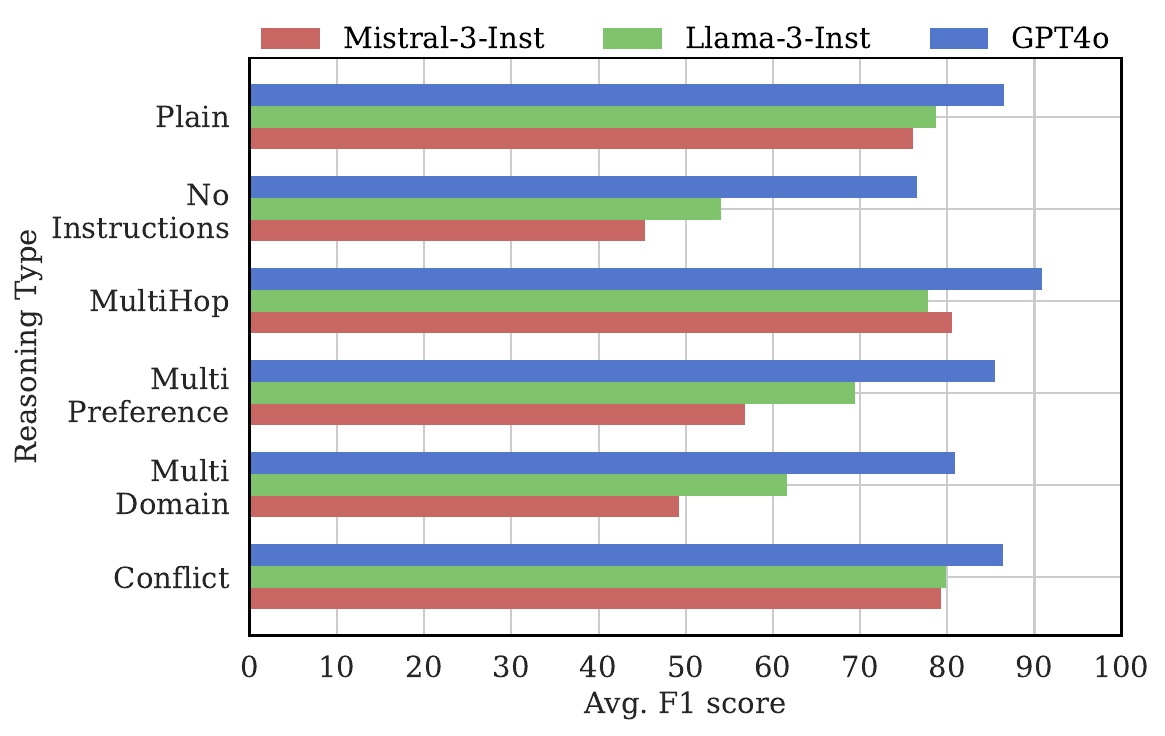}
        \caption{Average F1 scores of baseline models per each reasoning type. All scores are in \%.}
        \label{fig:example_type_scores}
\end{figure}

NLSI includes examples of varying difficulty based on the reasoning required to incorporate the standing instructions into the response (see Section 3.1. in \citet{moghe-etal-2024-interpreting} for a detailed description of the types). \autoref{fig:example_type_scores} demonstrates that while \gptb is able to consistently score above 75\% F1 on all reasoning types, open-source models fall behind. Both \mistralthreesin and \llamathreeschat can effectively follow simple, straightforward standing instructions where each argument of the final API call directly corresponds to one instruction (\textsc{Plain}, \textsc{Conflict}), suggesting some capability to solve the task. However, they struggle with cases that require reasoning across multiple domains (\textsc{MultiDomain}) or incorporating multiple preferences (\textsc{MultiPreference}). All models achieve lower scores when no instructions are provided (\textsc{NoInstructions}).

\subsubsection{Qualitative Analysis} \label{sec:qualitative}

\begin{figure}
    \centering
        \includegraphics[width=\linewidth]{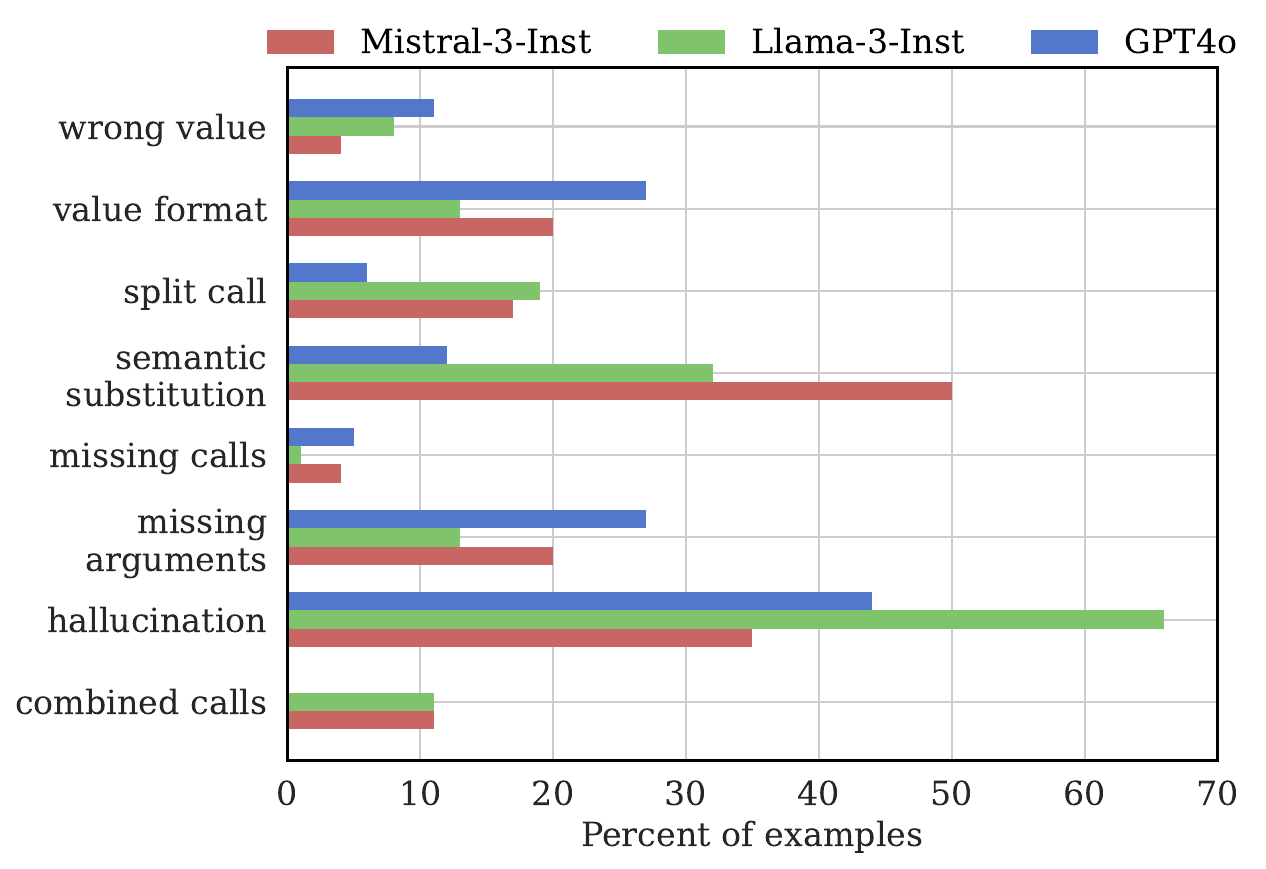}
        \caption{Distribution of errors on a sample of baseline predictions.}
        \label{fig:qual_analysis}
\end{figure}

Similarly to \citet{moghe-etal-2024-interpreting}, we manually annotate a sample of 100 predictions for each model and perform their qualitative analysis. We classify the errors into several categories (\autoref{tab:error_types} in \autoref{app:error-examples}) and present the results in \autoref{fig:qual_analysis}.

Open-source models frequently confuse semantically similar function and argument names (particularly \mistralthreesin, where the error is persistent on 50\% of the examples). This results in semantic substitution errors, where predictions are correct in meaning but deviate from documentation (e.g., using argument \texttt{city} from \texttt{GetRestaurants} instead of expected \texttt{location} in \texttt{GetTravel}). 35-75\% of examples include hallucinations, making it the most common error type for \llamathreeschat and \gptb. Hallucinations primarily involve the generation of extra arguments and the creation of new functions. We also observe value formatting issues, ranging from extracting part of the correct entity to canonicalisation issues, when models incorrectly unify date and time formats, which is common for \gptb (over 25\%). Often, LLMs ignore available information, missing one or several arguments, especially on examples requiring multi-hop reasoning (\textsc{MultiDomain}, \textsc{MultiPreference}). However, this happens in simpler cases as well (\textsc{Plain}, \textsc{Conflict}), where the models tend to favour one information source (the user query or instructions), leading to incomplete API calls. 

Overall, our findings support \citet{moghe-etal-2024-interpreting}. These results underline the task's inherent complexity and demonstrate that current LLMs cannot solve it on their own, highlighting the need for specialised methods to overcome this challenge.

\section{\name}
\begin{figure}[t]
    \centering
    \includegraphics[width=\linewidth]{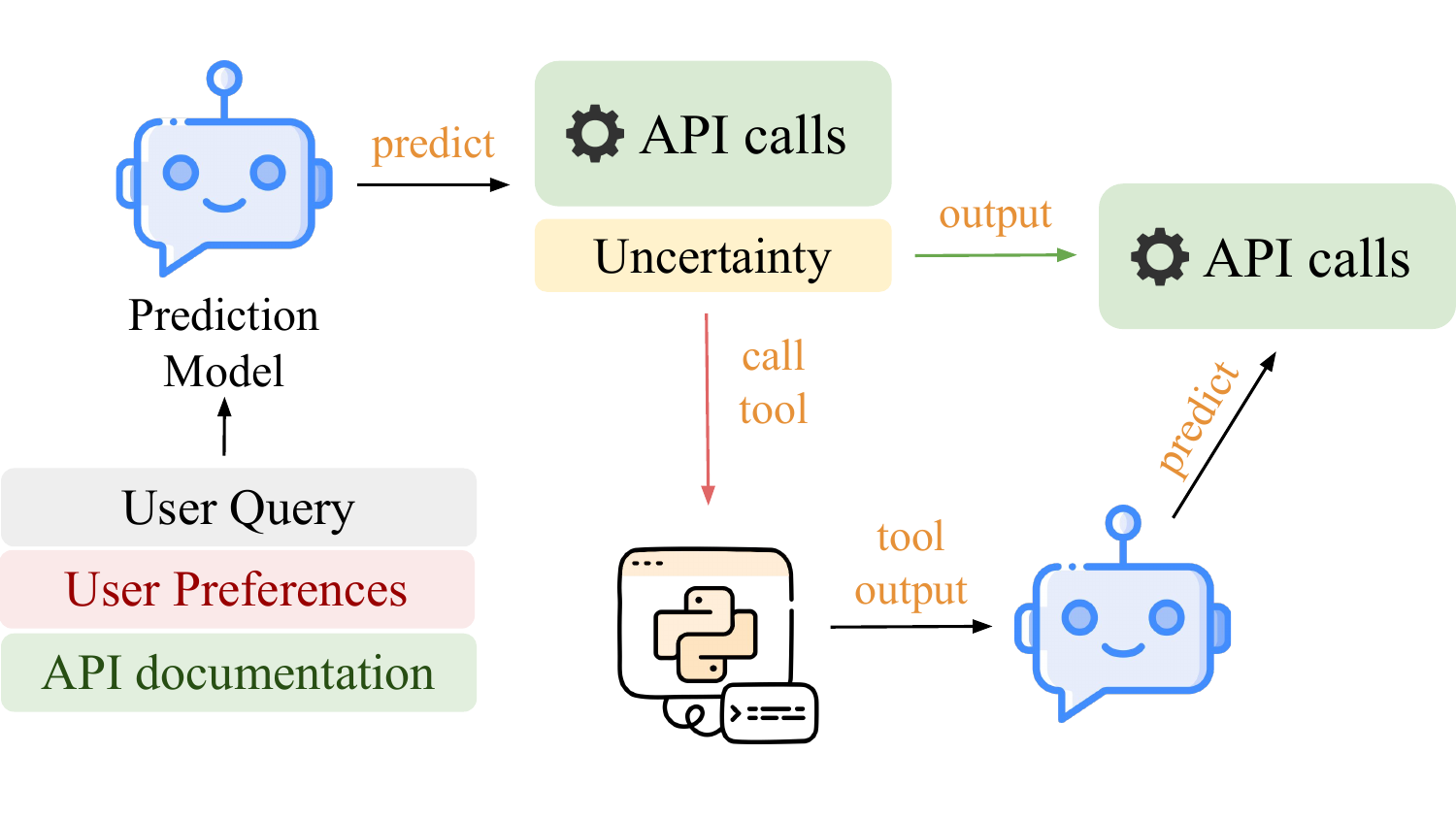}
    \caption{\name pipeline. An LLM first generates a response to the user query, and model uncertainty is extracted from its logits. Based on the uncertainty score, \name either accepts the response as is, or calls a structured tagging tool to augment the data before passing it back to the LLM and regenerating the answer.}
    \label{fig:final-pipeline}
\end{figure}

In this work, we aim to address key limitations of LLMs in personalised tool use, including semantic substitution errors, hallucinations, and missing arguments. We propose \name, a fully automated approach for goal-oriented dialogue that (i) employs a structured tagging tool for data augmentation and (ii) independently determines when tool use is required (iii) without additional training.  
\autoref{fig:final-pipeline} illustrates the full pipeline of \name, which we outline below. 

\subsection{Structured Tagging Tool} \label{sec:tools}

\begin{figure}[ht!]
    \centering
    \includegraphics[width=\linewidth]{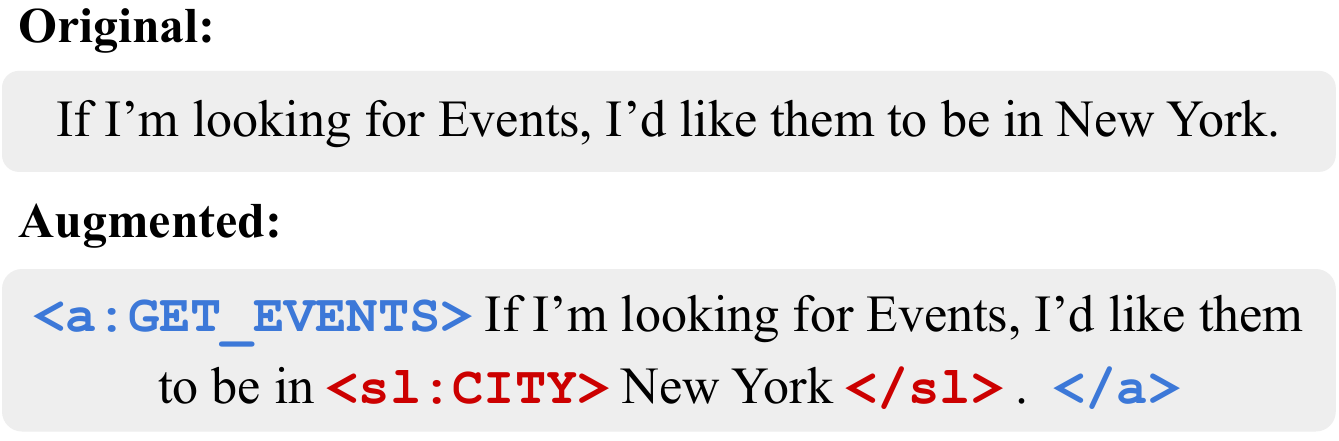}
    \caption{Example of structured tagging in \name. We use \texttt{\color{blue}<a:API> $\dots$ </a>} tags to denote relevant APIs and \texttt{\color{red}<sl:ARGUMENT> $\dots$ </sl>} to label arguments and their values.}
    \label{fig:data_aug_example}
\end{figure}

We define a data augmentation tool that introduces an intermediate representation between the natural language input and the function calls by annotating standing instructions with structured tags that encode action-level and slot-level information \autorefp{fig:data_aug_example}. Specifically, we label each instruction with hierarchical tags, where high-level action tags denote the relevant API and nested slot tags capture the arguments and their values. 
We call this approach \textbf{structured tagging}. Unlike traditional Named Entity Recognition or semantic parsing, which converts natural language to a structured representation, our method preserves the natural language aspect of instructions while introducing explicit nested tags, allowing models to leverage both the original instruction phrasing and explicit structural information. We hypothesise that adding this intermediate representation before code generation will facilitate more accurate API argument extraction and prevent information loss when generating API calls.

\newpage
Additionally, we explore two versions of the tool:

\begin{itemize}[leftmargin=*,topsep=0pt]
    \setlength\itemsep{0pt}
    \item \textbf{External Tagger} (\textbf{\extag}): Relies on an external model for tagging, allowing us to use specialised models with improved tagging accuracy. To isolate the effect of tagging quality, we use the same model for both tagging and subsequent API generation. Additionally, we include results where \gptb is used as an example of an optimal tagger (\textsc{\extag$_{\textsc{opt}}$}) to demonstrate how variations in tag quality influence overall task results (see \autoref{app:tagging} for tagger comparison).

    \item \textbf{\textsc{Tag-And-Generate}} (\textbf{\jointtag}): We ask the same base model to generate the augmentation for the standing instructions and the final API call jointly. This strategy allows us to rely on the internal reasoning abilities of an LLM, hypothetically making it easier for it to effectively use the provided information and predict the final answer.
\end{itemize}

\subsection{When to use a tool?}

Deciding when a tool is necessary is a challenging task. Recent approaches address tool detection through either an external learned classifier \cite{gemmell-dalton-2023-toolwriter} or reinforcement learning \cite{qiao-etal-2024-making}. Given the limited availability of training data for our task, we cannot rely on trainable methods. Thus, we propose to utilise model uncertainty to assess the confidence of an LLM in its prediction and determine whether additional help is needed to solve the task. 

We explore three methods for uncertainty estimation commonly used in text generation:

\begin{itemize}[leftmargin=*,topsep=0pt]
    \setlength\itemsep{0pt}
    \item \textbf{Sequence Margin}: the difference in the probability scores of the top two most likely predictions;
    \item \textbf{Margin@T}: the difference in the probability scores of the top $T$ most likely tokens, where $T$ is a hyper-parameter;
    \item \textbf{Least Confidence}: the difference between the probability of the top most confident prediction and 100\% confidence. The lower the score, the more certain the model is in its prediction.
\end{itemize}

\noindent
To choose the most effective method, we use the Pearson correlation coefficient \citep{freedman2007statistics} between the uncertainty of the model and the downstream task F1 metric on the validation set and report the results in \autoref{tab:pearson_corr} \autorefp{app:uncertainty}. 

Among the tested approaches, Least Confidence performs best, with a moderate correlation score (circa -0.45 for all models), suggesting that higher uncertainty indicates lower target scores. Other methods fail to provide reliable confidence estimates. Both weakly correlate with F1, making a comparison of top-2 most likely predictions, on sequence or token-level, unreliable. Thus, we choose Least Confidence as the main tool-use detector in \name.

To effectively utilise the uncertainty score, we select a threshold value on the validation set. The threshold is used to determine the confidence level of the model, based on which we choose to employ one of the following strategies: (i) output the model answer, or (ii) use a tool and regenerate the answer.

\section{Results \& Discussion} \label{sec:results}

In this section, we first investigate the effectiveness of \name's data augmentation tool on the NLSI task \autorefp{sec:data-aug}. Second, we illustrate the importance of tool detection and evaluate \name on the test subset in NLSI \autorefp{sec:overall-res}. Finally, we perform behavioural analysis of \name's predictions when both structural tagging and tool detection are utilised to demonstrate the impact of the approach \autorefp{sec:pred_analysis}.

\subsection{Effects of Structured Tagging} \label{sec:data-aug}
\begin{table}[t!]
    \centering
    \resizebox{\linewidth}{!}{%
    \begin{tabular}{llll}
    \toprule
        \textbf{Config Name} & \textbf{Tags} & \textbf{Tagger Model} & \textbf{Tool Detector} \\
    \midrule
        \textsc{Default} & \xmark & \xmark & \xmark \\
        \extag & \checkmark\xspace External & Base model & \xmark\xspace Naïve \\
        \extag$_{\textsc{opt}}$ & \checkmark External & \gptb & \xmark\xspace  Naïve \\
        \jointtag & \checkmark\xspace Joint & Base model & \xmark\xspace Naïve\\
        TAPS & \checkmark\xspace External & Base model & \checkmark\xspace Uncertainty\\\xspace
        TAPS$_{\textsc{opt}}$ & \checkmark\xspace External & \gptb & \checkmark\xspace Uncertainty\\
        TAPS-\textsc{Oracle} & \checkmark\xspace External & Base model & \checkmark\xspace Oracle\\
        TAPS-\textsc{Oracle$_{\textsc{opt}}$} & \checkmark\xspace External & \gptb & \checkmark\xspace Oracle \\
    \bottomrule
    \end{tabular}}
    \caption{Model configurations used in experiments.}
    \label{tab:configs}
\end{table}

To show the effectiveness of structured tagging, we compare the performance of both tagging tools to default models without tools. We naïvely apply the tool to all instances in the validation set. Here and in further experiments, we use ICL to evaluate the models and optimise model performance by bootstrapping a set of demonstrations with random search \citep{khattab2023dspy}. \autoref{tab:configs} summarises all model configurations used in our experiments.
Full implementation details are in \autoref{app:exp-details}.

\begin{table}[t!]
\centering
    \resizebox{\linewidth}{!}{
    \begin{tabular}{llcccc}
    \toprule
        \textbf{Model} & \textbf{Aug.} & \textbf{EM} $\uparrow$ & \textbf{F1} $\uparrow$ & \textbf{Prec.} $\uparrow$ & \textbf{Rec.} $\uparrow$ \\
        \midrule
        \llamathreeschat & \textsc{Default} & 42.23 & 78.19 & 80.30 & 78.60 \\
        & \textsc{\extag} &  \underline{44.22} &  \underline{80.34} &  81.90 &  \underline{81.49} \\
        & \textsc{\extag$_{\textsc{opt}}$} & \textbf{51.79} & \textbf{84.46} & \textbf{86.39} & \textbf{84.86} \\
        & \textsc{\jointtag} & 41.43 & 78.31 & \underline{82.97} & 77.19 \\
        \midrule
        
        \mistralthreesin & \textsc{Default} & 30.68 & 64.21 & 65.21 & 65.37 \\
        & \textsc{\extag} & \underline{36.65} & \underline{75.12} & \underline{78.33} & \underline{74.30} \\
        & \textsc{\extag$_{\textsc{opt}}$} & \textbf{42.63} & \textbf{79.34} & \textbf{82.23} & \textbf{79.04} \\
        & \textsc{\jointtag} & 33.47 & 66.66 & 70.56 & 64.97  \\

        \midrule
        \gptb & \textsc{Default} & \underline{56.18} & \underline{87.40} & \textbf{90.41} & \textbf{86.83} \\
        & \textsc{\extag} & \textbf{57.37} & \textbf{87.47} & \underline{89.63} & \underline{86.72} \\
        & \textsc{\jointtag} &  52.99 & 83.94 & 86.00 & 83.24 \\

         \bottomrule
         
    \end{tabular}}
    \caption{Model performance with and without naïve tool-use.   \textsc{\extag}: the same model is used for tagging and API call generation sequentially; \textsc{\extag$_{\textsc{opt}}$}: tagging is performed by a separate, high-performing tagger; \textsc{\jointtag}: tags and API calls are generated jointly in a single step. \textbf{EM}: exact match. \textbf{F1}: Slot-wise F1 score. \textbf{Prec.}: precision. \textbf{Rec.}: recall. All scores are in \%. Best performance is in \textbf{bold}, second best is \underline{underlined}.}
    \label{tab:data_aug_effects}
\end{table}

 We report the results in \autoref{tab:data_aug_effects}. We observe marginal gains in GPT-4o when using \extag, and consistent improvements across all four metrics for open-source models: \llamathreeschat and \mistralthreesin improve EM by 2\% and 6\%, respectively, and up to 11.9\% when the optimal model is used for tagging \extag$_{\textsc{opt}}$).

We further investigate the impact of tool use on model outputs and calculate the percentage of predictions that improve or degrade after structured tagging is applied \autorefp{fig:tool-errors}. Overall, all models benefit from tool use in less than 50\% of cases, with open-source models gaining the most.
Only 16.3\% of predictions improve for \gptb, which is least affected by tagging, with more than 62\% of predictions remaining the same with and without the tags, compared to 37\% for both open-source models. In 16-27\% of cases, LLMs score lower when having the tags. Below, we discuss our key findings regarding structural tagging effects.

\begin{table}[th]
    \resizebox{\linewidth}{!}{\centering
    \begin{tabular}{lccc}
       \toprule
       \textbf{Result} & \textbf{\llamathreeschat} &\textbf{\mistralthreesin} & \textbf{\gptb}\\
       \midrule
       Win $\uparrow$ &  35.1 & 45.8 & 16.3 \\
       Same &  37.0 & 37.9 & 62.2\\
       Loss $\downarrow$ & 27.9 & 16.3 & 21.5\\
       \bottomrule
    \end{tabular}}
    \caption{Data augmentation effects for \extag$_{\textsc{opt}}$. All scores represent \% of instances. All calculations are based on F1.}
    \label{fig:tool-errors}
\end{table}

\paragraph{LLMs struggle to map natural language to code.} 
The inferior performance of all \textsc{Default} models compared to \extag\xspace suggests that LLMs still need additional tools to successfully generate code from natural language when complex reasoning is required.
Strong results of \extag, even with lower quality tags, support our hypothesis that introducing an intermediate representation between natural language and code can significantly enhance model performance. Notably, tagging is less effective for \gptb. We hypothesise that this can be due to the \gptb's stronger in-context reasoning, making task decomposition less beneficial, compared to smaller models that generally do not perform complex tasks as well. The improvements of \extag$_{\textsc{opt}}$ over \extag\xspace show that the effectiveness of the proposed approach is closely tied to the reliability of structured tags. Initial robustness experiments confirm this sensitivity, and we further investigate the impact of tag quality in \autoref{sec:robust}. 

\paragraph{Internal reasoning does not boost the interpretational abilities of LLMs.} Our results demonstrate that explicitly prompting the models to generate structured tags before producing the function calls is not uniformly effective. The observed decrease in recall suggests that this approach may result in some information loss. While LLMs generate more accurate code, they tend to omit more arguments, showing that solving the task end-to-end is difficult. An additional explanation for such behaviour is the demonstration optimisation strategy we use. Existing ICL optimisation approaches do not support optimisation for multiple outputs, leading to suboptimal model performance.

\paragraph{Naïve tool use fails to yield consistent improvements.} We show that naïvely leveraging the tool is inefficient, both in compute and target metrics and sometimes even counterproductive. This highlights the importance of tool detection to determine if a tool is required on the instance level.

\subsection{Tool Detection Effects} \label{sec:overall-res}

We evaluate \name on the NLSI test set and report the results in \autoref{tab:final-results}. We compare the scores with lower-boundary baselines, default models without tools and naïve tool use (\extag), and upper-boundary oracle models optimised for tool detection. 
The oracle prediction is compiled by retrospectively selecting the examples that actively benefit from tool use and leaving other predictions unchanged.

\begin{table}[t!]
\centering
\resizebox{\linewidth}{!}{
    \begin{tabular}{llcccc}
    \toprule
       \textbf{Model} & \textbf{Config} & \textbf{EM} $\uparrow$ & \textbf{F1} $\uparrow$ 
       & \textbf{Prec.} $\uparrow$ & \textbf{Rec.} $\uparrow$ \\
    \midrule

        \llamathreeschat    & \textsc{Default} &  41.76 & 78.26 & 82.96 & 76.80 \\
                            & \extag & 47.55 & 82.28 & 84.89 & 81.88 \\
                            & \textsc{\name } & \underline{51.18} & \underline{83.94} & \underline{87.20} & \underline{82.95} \\
                            & \textsc{\name-Oracle} & \textbf{59.07} & \textbf{88.42 }& \textbf{91.60 }& \textbf{ 87.11} \\
                            \noalign{\vskip 0.5ex}\cdashline{2-6}\noalign{\vskip 0.5ex}
                            & \extag$_{\textsc{opt}}$ & 51.23 & 84.51 & 87.23 & 83.86 \\
                            & \textsc{\name$_{\textsc{opt}}$} & \underline{53.04} & \underline{85.64} & \underline{88.67} & \underline{84.56} \\
                            & \textsc{\name-Oracle$_{\textsc{opt}}$} & \textbf{59.85} & \textbf{89.65} & \textbf{92.82} & \textbf{88.10} \\

        \vspace{-1em}\\
        \midrule
        \vspace{-1em}\\

        \mistralthreesin    & \textsc{Default} & 35.74 & 69.11 & 70.64 & 69.83 \\
                            & \extag & 40.29 & 76.20 & 79.53 & 75.57 \\
                            & \textsc{\name} & \underline{42.40} & \underline{76.76} & \underline{79.74} & \underline{76.32} \\
                            & \textsc{\name-Oracle} & \textbf{48.82} & \textbf{81.79} & \textbf{84.29} & \textbf{81.33} \\
                            \noalign{\vskip 0.5ex}\cdashline{2-6}\noalign{\vskip 0.5ex}
                            & \extag$_{\textsc{opt}}$ & 42.35 & 78.55 & 82.63 & 77.24 \\
                            & \textsc{\name$_{\textsc{opt}}$} & \underline{44.17} & \underline{79.03} & \underline{82.66} & \underline{78.04} \\
                            & \textsc{\name-Oracle$_{\textsc{opt}}$} & \textbf{49.85} & \textbf{83.19} & \textbf{86.19} & \textbf{82.36} \\
        
        \vspace{-1em}\\
        \midrule
        \vspace{-1em}\\

        \gptb   & \textsc{Default} & 56.32 & 86.99 & 89.25 & 86.91 \\
                & \extag & 55.54 & 86.49 & 88.78 & 85.65 \\
                & \textsc{\name} &  \underline{58.63} & \underline{87.86} & \underline{90.03} & \underline{87.21} \\
                & \textsc{\name-Oracle } & \textbf{65.88} & \textbf{91.46} & \textbf{93.57} & \textbf{90.49}\\

         \bottomrule
    \end{tabular}}
    \caption{Model performance on test data. \textsc{opt}: best performing model is used for tagging. \textbf{EM}: exact match. \textbf{F1}: Slot-wise F1 score. \textbf{Prec.}: precision. \textbf{Rec.}: recall. All scores are in \%. Best performance is in \textbf{bold}, second best is \underline{underlined}.}    \label{tab:final-results}
\end{table}

Overall, we find that for open-source models, both naïve tool use and \name are superior to base models without tools by a margin with EM and F1 gains of up to 10\% when an optimal tagger is employed. Using a tool detector significantly improves target metrics compared to naïve tool use, with \name and \name-\textsc{Oracle} outperforming \extag\xspace by 2/8\% EM, respectively. The results for \gptb are less consistent, however, they illustrate the same idea. While \extag\xspace leads to model score degradation, leveraging a tool detector improves model effectiveness by 2/9\% EM. Although optimal tags significantly outperform lower-quality ones in the naïve setting, tool detection narrows the gap to within 1\% EM, demonstrating that the approach is effective even when a lightweight model is used for all steps of the pipeline. We highlight our key findings below.

\paragraph{Using a tool detector can maximise tool use effectiveness. }
We find that both \name and \name-\textsc{Oracle} outperform all baseline models, demonstrating that selectively using tools is much more effective than relying on them at all times. Moreover, our experiments show that tool detection allows us to minimise both time and compute spent on the task by applying the tool 20\% fewer times for open-source models and over 55\% fewer times for \gptb when using uncertainty, and up to 80\% in the oracle case. This is particularly valuable, as achieving an optimal balance between latency and model capabilities is crucial for task assistants interacting with users in real time. 

\paragraph{Using uncertainty for tool detection is possible but suboptimal}
While we demonstrate that utilising uncertainty for tool detection can be beneficial, we note the suboptimal performance of \name compared to the oracle model. \name-\textsc{Oracle} is consistently superior to \name for all models, with performance gains of 5.7-7.9\% w.r.t. EM scores. The same trend is observed in terms of resource efficiency. This indicates that uncertainty may not be the most effective approach to determine whether calling a tool would yield higher scores, and alternative methods may be explored in future.

\begin{figure}[t]
    \centering
    \includegraphics[width=\linewidth]{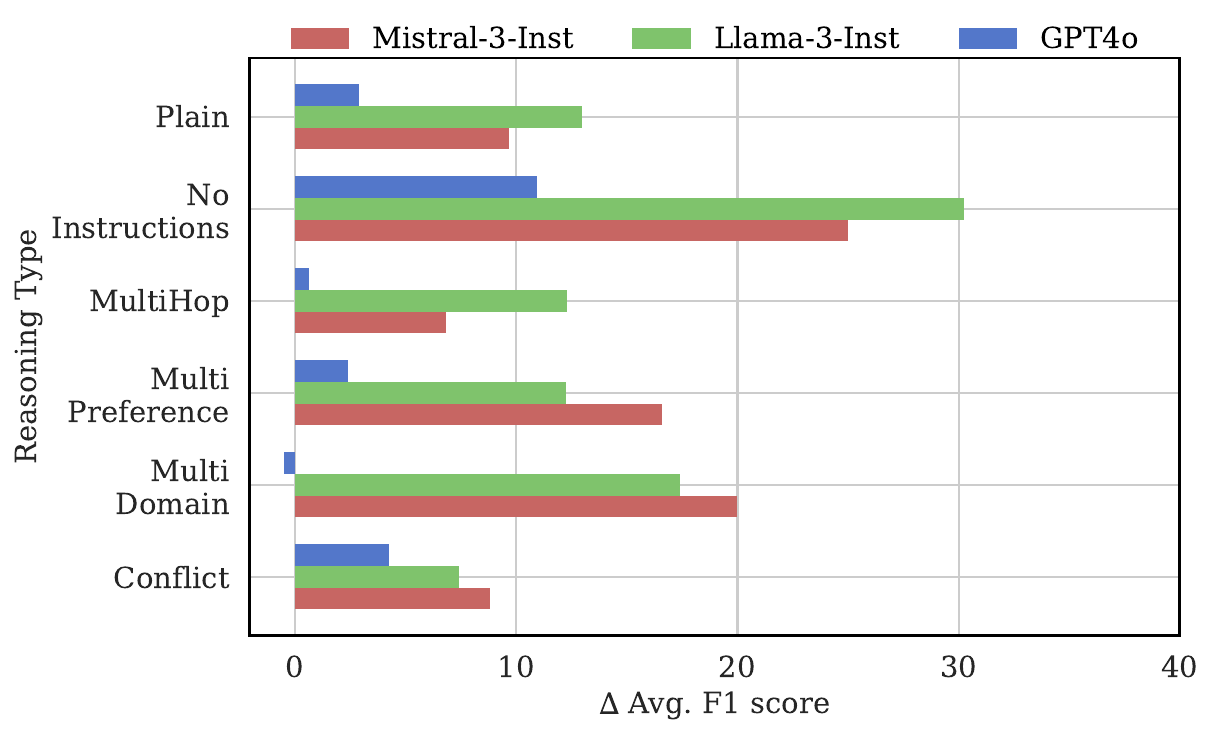}
        \caption{$\Delta$ F1 scores of \name models compared to baselines per each reasoning type.}
\label{fig:example_type_scores_delta}
\end{figure}

\subsection{Prediction analysis} \label{sec:pred_analysis}

\autoref{fig:example_type_scores_delta} demonstrates the difference in F1 scores of baseline models \autorefp{sec:behaviour} and \name (for avg. F1 scores refer to \autoref{app:extra-results}). We observe consistent improvements in scores or on-par performance when using \name on all reasoning types. An external tool for tagging increases the performance by up to 30\% (\llamathreeschat) on the task, with an average improvement on each reasoning type by 3-15\% depending on the model. 

\begin{figure}[t]
        \centering
    \includegraphics[width=\linewidth]{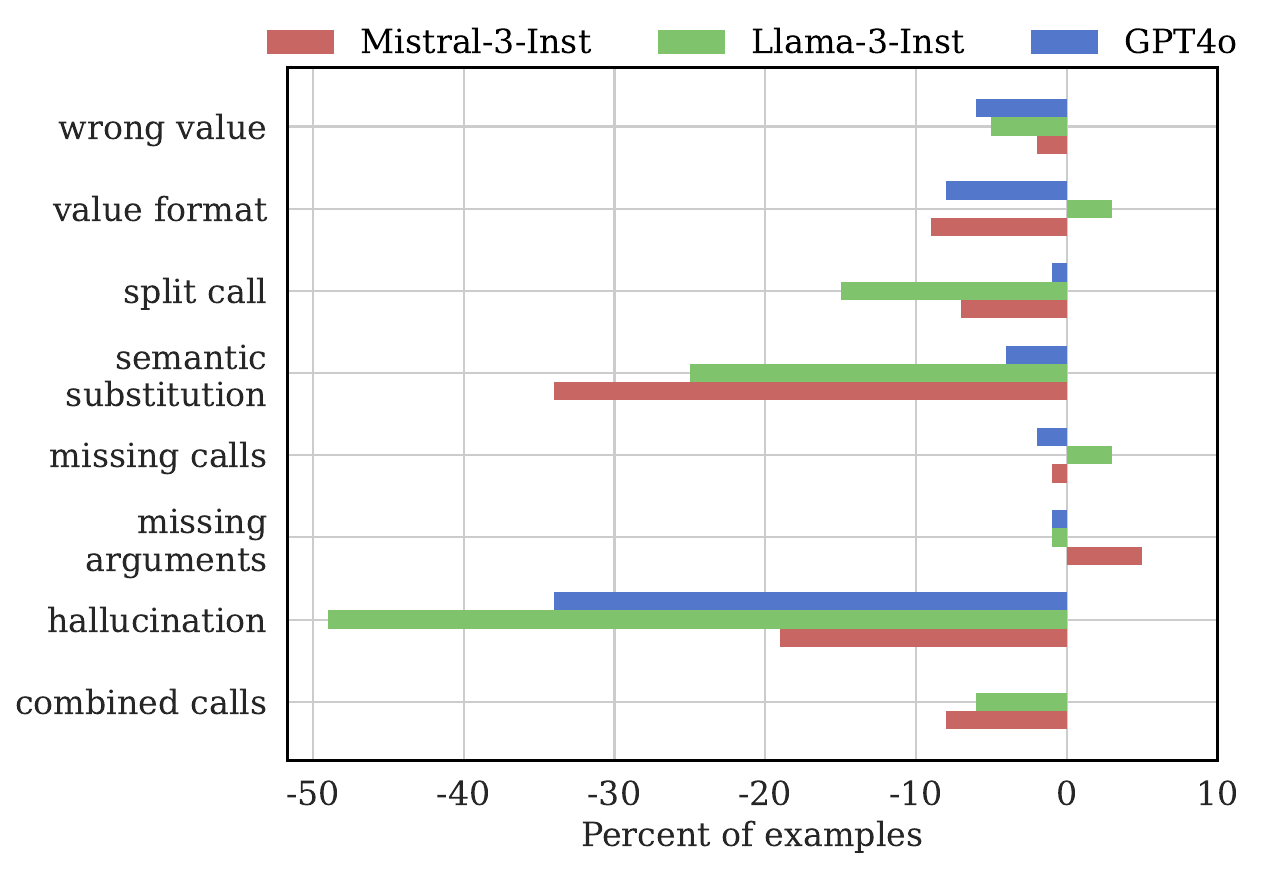}
    \caption{Changes in the distribution of errors in \name compared to baseline models. The scores represent the percentage of examples that improved or degraded with \name.}
    \label{fig:qual_analysis_delta}
    \end{figure}

We sample and manually annotate the same 100 examples for each model as in \autoref{sec:qualitative} and compare the percentage of errors. \autoref{fig:qual_analysis_delta} presents the results of the comparison. The biggest difference is observed on hallucinations (19-49\% less errors) and semantic substitution errors (4-34\% decrease), which we specifically targeted with our approach. However, we also notice slight increases in some error types for some models, which can be due to error propagation since we are using \gptb as a tagger model. For example, \llamathreeschat exhibits more value formatting issues when using a data augmentation tool, one of the common errors of \gptb, according to our baseline evaluation (\autoref{fig:qual_analysis}). Additionally, we notice an increase in missing arguments for \mistralthreesin, specifically when shared contextual information (e.g. location) is available. We attribute this to our tagging approach, which does not allow us to incorporate the links between instructions, leading to the exclusion of some possible annotations. We discuss this limitation in more detail in \autoref{sec:limitations}. Overall, we show that using \name significantly decreases the number of errors for all models, proving it an effective solution for tool use personalisation.

\section{Tagging Sensitivity Analysis} \label{sec:robust}

While our main experiments \autorefp{sec:results} demonstrate that structured tagging substantially enhances the ability of LLMs to incorporate user preferences in tool calling, they also reveal that the effectiveness of this approach varies with the quality of the tags. This raises a question of the robustness of LLMs to imperfect and noisy tags. To address this, we conduct a controlled corruption study where we systematically perturb the tags provided to the model. Starting from golden tags, manually annotated by one of the authors, we randomly corrupt $n\%$ of them, with $n \in \{0, 10, ..., 100\}$. Possible tag corruptions are sampled from error types that frequently occur in real tagger outputs, including slot deletion (replicating missing arguments error), tag boundary shifts, and semantic substitution of slot and function names. The corrupted tags are then passed to the \extag\xspace pipeline, and the final model predictions are evaluated using the same setup as in the main experiments. This allows us to quantify the sensitivity of different models to varying tag quality and to identify which models are more robust to noisy annotations.

\begin{figure}[t!]
    \centering
    \includegraphics[width=\linewidth]{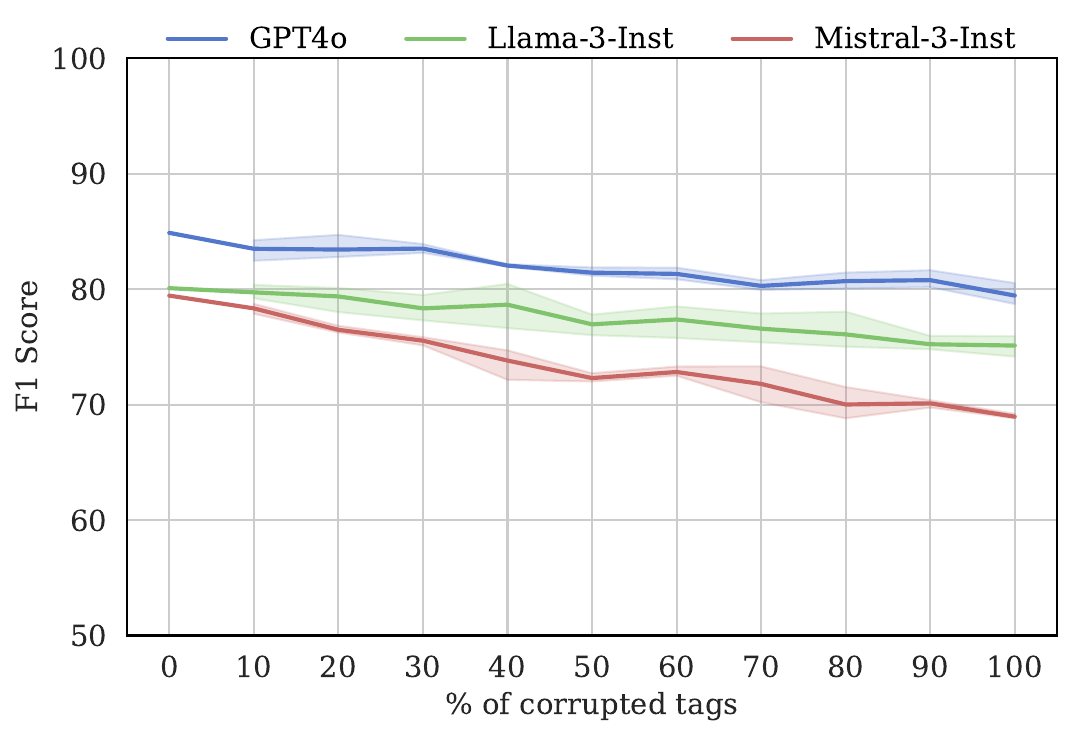}
    \caption{The effects of tag quality on the F1 scores of \gptb, \llamathreeschat, and \mistralthreesin. All scores are in \%.}
    \label{fig:tag_ablation_f1}
\end{figure}

\autoref{fig:tag_ablation_f1} presents the results of the study with respect to the F1 scores, other metrics follow the same trend (see \autoref{fig:tag_effects}). Across all models, the scores degrade monotonically as the percentage of corrupted tags increases. This confirms that the noise in the structured tags directly impacts the downstream performance of LLMs. However, the sensitivity to tag quality differs between the models. \gptb and \llamathreeschat remain relatively robust, with the tag quality degrading by up to 5\%. \mistralthreesin exhibits a steeper decline, with F1 scores dropping more than 10\% if comparing the 100\% corruption rate and golden tags. These results align well with our findings in \autoref{sec:data-aug}, showing that higher-quality tags (marked with \textsc{OPT}) yield more drastic improvements compared to suboptimal ones. Relative to the no-tag baseline (\textsc{Default}, \autoref{tab:data_aug_effects}), we find that structured tagging improves models' scores when tag quality is sufficiently high, but harms them beyond a certain level of corruption, showing that overly noisy tags can reduce downstream effectiveness.

Overall, this analysis shows that the effectiveness of \extag\xspace directly depends on the quality of the produced tags. Together with the TAPS results, these findings demonstrate both the potential and the limitations of leveraging structured tags. While they can significantly improve the ability of LLMs to produce accurate personalised tool calls, their effectiveness is bound by upstream tag quality and model robustness. This suggests two promising directions for future work: (i) developing more accurate taggers, and (ii) designing approaches that explicitly account for and compensate for noisy intermediate tags. More broadly, these results highlight the need for better entity understanding in LLMs, which remains one of the key bottlenecks for robust personalised tool use.

\section{Related Work}
\paragraph{Tool-Augmented Language Models} Introduction of tool-augmented LLMs have enabled general agents to perform a variety of diverse tasks \citep{parisi2022talmtoolaugmentedlanguage, patil2023gorillalargelanguagemodel, mialon2023gaiabenchmarkgeneralai}. A body of work on tool use leverages the innate abilities of LLMs to produce structured data from natural language input \citep{song2023restgptconnectinglargelanguage, liu2023controlllmaugmentlanguagemodels, liu2024toolnetconnectinglargelanguage, zhang-etal-2024-reverse}. For example, \citet{hsieh2023tooldocumentationenableszeroshot} show that tool documentation alone is sufficient to elicit tool use in LLMs without demonstrations. Some use task decomposition \citep{wu2024sealtoolsselfinstructtoollearning} and a backward reasoning pipeline \citep{zhang-etal-2024-reverse} to generate appropriate parameter values effectively.
Other works incorporate tuning-based approaches \citep{parisi2022talmtoolaugmentedlanguage, schick2023toolformerlanguagemodelsteach, patil2023gorillalargelanguagemodel, mekala2024toolverifiergeneralizationnewtools, shen2024smallllmsweaktool}, with \citet{shi2024learningusetoolscooperative} iteratively predicting and filtering tool-usage plans, and \citet{qiao-etal-2024-making} leveraging reinforcement learning with tool execution feedback for consistent tool invocation. \citet{hao2023toolkengpt} train tool embeddings, while \citet{shen2024smallllmsweaktool} propose a two-stage fine-tuning technique with join training and separate refinement of specialised modules for each subtask in tool-use paradigm. Despite their effectiveness, existing TALMs still face challenges in personalising interactions and efficiently integrating tool use with conversational history.

\paragraph{Personalisation}
Personalisation is an important aspect of any system interacting with users. Many works on personalisation for dialogue provide models with user profiles, describing their preferences and personality traits through natural language statements \citep{li-etal-2016-persona, zhang-etal-2018-personalizing, majumder-etal-2020-like} or structured databases (\citealt[among others]{song-etal-2020-profile, aliannejadi2024ikat}). \citet{cheng-etal-2024-dialogues} propose to learn user preferences from dialogue history. Nevertheless, these works focus on creating a user persona for more engaging conversations rather than task completion. \citet{joshi2017personalization} introduce simple structured user profiles for a limited number of goal-oriented dialogue tasks and explore rule-based systems and memory networks. To the best of our knowledge, \citet{moghe-etal-2024-interpreting} is one of the only approaches that attempts to personalise goal-oriented dialogue through explicit and complex user preferences in natural language. However, the work explores only simple ICL approaches for the task. Our work attempts to solve the task by leveraging tool use and an internal tool detection mechanism that provides more flexibility and robustness in tailoring tool use according to user preferences.

\section{Conclusion and Future Work}

In this work, we explore the limitations of LLMs to perform the personalised tool use task. We find that all LLMs struggle to effectively incorporate user preferences, especially when complex reasoning is required, suffering from semantic errors, information loss and hallucinations. To combat this, we propose \name, a tuning-free solution for personalised tool use in task assistants. \name combines (i) a structural tagging tool that introduces an intermediate representation between natural language and code and (ii) an internal tool detector to facilitate the incorporation of user preferences for tool use in goal-oriented dialogue. We conduct a thorough analysis of widely used LLMs on the NLSI dataset and demonstrate that our method consistently outperforms pre-trained open-source models of the same size. We show that \name enables the models to more effectively reason and infer tool calls from user queries and successfully incorporate information from personalised user preferences, all while being fully automatic and not requiring additional training. Through ablation studies, we show that each component in \name plays an important role in the solution of the task, significantly minimising most error types for tested LLMs. We hope our work will inspire more research on incorporating extended context in tool use in future.

\section*{Limitations} \label{sec:limitations}
\paragraph{A better structural tagger is required.} One of the limitations of our solution lies in the tagging approach we employ, which has several shortcomings. First, as briefly mentioned in \autoref{sec:pred_analysis}, we label APIs and arguments on the sentence level only and do not consider the whole user profile. This leads to the loss of shared contextual information, which should be included in all relevant API calls but is tagged as belonging to only one API.  
Second, we apply the tool only to the user profile, which might lead to some information loss, as we do not explicitly label the relevant information from user queries, prompting the model to prioritise user profiles over queries. Lastly, in our experiments, we use ICL and prompting, while training a specialised model for tagging might yield better and more reliable results. A more sophisticated tagging procedure will help mitigate those issues, and we hope to continue working in this direction in future.

\paragraph{LLMs are not robust to changes in input.} We utilise LLMs' in-context learning abilities to create a solution for the task. Such an approach is less computationally expensive, as it does not require additional training and allows for generalisation to unseen domains, functions and tasks. However, we do not address a well-known shortcoming of ICL, namely its sensitivity to prompt template choice and demonstration selection \citep{lu-etal-2022-fantastically, chang-jia-2023-data, sclar2024quantifyinglanguagemodelssensitivity}. While we explore several prompts in our preliminary studies and utilise demonstration optimisation, we do not conduct extensive experimentation on the topic as it is not the primary focus of our work. This means that the prompts used to evaluate \name may not be optimal for the task. While training a specialised model for the task would seem like a logical solution, the dataset size is insufficient for straightforward fine-tuning and requires a different approach. For example, LIMA \citep{zhou2024lima} or similar methods can be used to fine-tune a model on low data cases. 

\paragraph{The need for a better evaluation benchmark.} In our experiments, we use the NLSI dataset, collected by \citet{moghe-etal-2024-interpreting}, as the only dataset, to our knowledge, that incorporates user preferences into tool-augmented conversational agents. However, the dataset has several downsides. First, the dataset is created automatically from templates without additional validation, so it contains some errors (see \autoref{sec:qualitative}) and is overall not as diverse and natural in terms of both language and domains covered. Additionally, evaluation on NLSI is based on comparing code strings rather than the actual tool output. This approach can underestimate model performance, as two different programs can lead to the same output when executed but will get different evaluation scores. 
Therefore, we acknowledge the need for a better evaluation methodology and benchmark for the task in order to more accurately assess and compare the capabilities of LLMs with respect to contextualised tool use. 

\section*{Ethical Considerations}\label{sec:ethics}
\noindent Privacy is a critical concern in natural language processing, especially when handling personal data \citep{horvitz2015data, yao2024survey-privacy, miranda2025preserving}. Working with user preferences and extended dialogue history can inadvertently lead to the potential exposure of sensitive personal information. Our approach employs in-context learning, which prevents the model from memorising private information. This strategy aligns with the growing emphasis on privacy in LLMs by ensuring that user data remains protected throughout the conversation.

We improve and proofread the text of this paper using Grammarly\footnote{\href{https://github.com/huggingface/accelerate}{grammarly.com}} to correct
grammatical, spelling, and style errors and paraphrasing sentences.
\section*{Acknowledgements}
\noindent This work is supported by a Turing AI Acceleration Fellowship from the Engineering and Physical Sciences Research Council, grant number EP/V025708/1.
\bibliography{anthology,custom}

\appendix
\clearpage
\newpage
\section{Experiment Details}\label{app:exp-details}
\subsection{Dataset Statistics}
\noindent We run all of the experiments of NLSI \citep{moghe-etal-2024-interpreting}, which has a train/validation/test splits of sizes 150/251/2040 instances. We refer you to the original paper for full details on the data.

\subsection{Baseline Evaluation \autorefp{sec:behaviour}} 
\noindent For baseline evaluation, we use the prompt, provided by \citet{moghe-etal-2024-interpreting} for all our models \autopromptref{app:baseline-prompt}. We set the number of few-shot demonstrations to 1 and use default model parameters.

\subsection{Main Experimental Settings (\autoref{sec:results})} 
\paragraph{Optimiser settings} For all experiments in \name we optimise the ICL examples using BootstrapFewShotWithRandomSearch algorithm \citep{khattab2023dspy}. We set the following parameters to the optimiser: 

\begin{itemize}[topsep=5pt]
    \setlength\itemsep{0pt}
    \item \texttt{max\_bootstrapped\_demos} = 1 for \gptb and \llamathreeschat in the {\jointtag} setting else 5
    \item \texttt{max\_labeled\_demos} = 5
    \item \texttt{num\_candidate\_programs} = 5 (\gptb) / 10 (other models)
    \item \texttt{num\_threads} = 1
    \item \texttt{metric} = {\dbq}exact\_match\dbq
\end{itemize}

\paragraph{Prompt Selection}
We conduct a simple prompt selection experiment on the validation set of NLSI and choose the following prompts for our main experiments with \name. To evaluate all LLMs in \textsc{Default} setting, we use \autopromptref{app:def-prompt} for \llamathreeschat and \autopromptref{app:def-prompt-2} for \mistralthreesin and \gptb. For \textsc{\extag} we select \autopromptref{app:tag-simple-prompt} for \llamathreeschat and \gptb and \autopromptref{app:tag-simple-prompt-2} for \mistralthreesin. All runs in \textsc{Joint-Tag} configuration use \autopromptref{app:tag-gen-prompt} as the prompt.

\paragraph{Generation Parameters}
To select the optimal generation parameters for \mistralthreesin and \llamathreeschat models, we run a simple grid search on the validation set. For all our experiments we use the default set of generation parameters for \gptb and the following for open-source models (when different parameters for \mistralthreesin and \llamathreeschat are used, we report them with a forward-slash):

\begin{itemize}[topsep=5pt]
    \setlength\itemsep{0pt}
    \item \texttt{num\_beams} = 5 / 2
    \item \texttt{do\_sample} = True 
    \item \texttt{temperature} = 0.85 / 0.95
    \item \texttt{top\_k} = 50
    \item \texttt{top\_p} = 1.0
\end{itemize}

\paragraph{Tool Detection Parameters} 
We use Least Confidence as our main tool detection strategy for all the experiments. We select the threshold for each model on the validation set. The following threshold values are used: 0.02 (\llamathreeschat), 0.01 (\mistralthreesin), and 0.04 (\gptb).

\paragraph{GPU-Usage} We use one 40GB A100 GPU, setting the batch size of 1. It takes approximately 1.5-5 hours to run one experiment on the whole validation set and 5-13 hours to make a full pass over the test set depending on the model and generation parameters.

\section{Selection of the Tagger Model}\label{app:tagging}
To choose the models for the \extag\xspace strategy, we manually annotate the validation subset of data and compare automatically generated tags with the golden standard. To assess the tagger models we treat the task as a standard token classification problem and calculate macro-averaged F1, precision, and recall. We use \autopromptref{app:data-aug-tagger} for all models to generate tags for standing instructions and set all generation parameters to default values. All models are assessed in one-shot configuration. We do not optimise the demonstrations, but use a static example created manually for all instances. The results of the evaluation are presented in \autoref{tab:tag_aug_model_eval}.

\begin{table}[hb!]
\centering
    \begin{tabular}{lccc}
    \toprule
    \textbf{Model}   & \textbf{F1 $\uparrow$} & \textbf{Prec. $\uparrow$} & \textbf{Rec. $\uparrow$} \\ \midrule
      \codellamasin & 63.98 & 63.09 & 65.89 \\
      \llamatwoschat & 63.18 & 62.75 & 63.97 \\
      \llamathreeschat  & 71.54 & 74.82 & 69.34  \\
      \mistralthreesin  & 76.72 & 77.04 & 77.12 \\
      \gptb & 87.18 & 87.00 & 87.44 \\

      \bottomrule
    \end{tabular}
    \caption{Tagging performance on the manually annotated validation set.  \textbf{F1}: macro-average F1 score. \textbf{Prec.}: precision. \textbf{Rec.}: recall. The best result is in \textbf{bold}, second best is \underline{underlined}. All scores are in \%.}
    \label{tab:tag_aug_model_eval}
\end{table}

\paragraph{Results} 
Our experiments show, that open-source LMs are still far behind \gptb when it comes to their ability to augment input with tags. While \gptb scores exceed 86\%, the difference between the best-performing open-source LLM (\mistralthreesin) and \gptb reaches 10\%. Despite being the only model trained specifically to handle code and structured data, \codellamasin yields one of the lowest scores on the task with F1 of 63\%. Despite \gptb outperforming all open-source LLMs in the task, its performance is still does not exceed 90\%, leaving room for improvement. We acknowledge this but continue to use \gptb as our main external tagger model for \extag.

\section{Uncertainty Estimation}\label{app:uncertainty}
\begin{table}[ht]
    \centering
    \begin{tabular}{lc}
    \toprule
        \textbf{Method} & \textbf{Statistic} \\
    \midrule
      Least Confidence &  \textbf{\textit{-0.452}} \\
      Margin@1 & 0.145 \\
      Margin@2 & \textit{0.317} \\
      Margin@3 & \textit{0.314} \\
      Margin@4 & \textit{0.295} \\
      Margin@5 & \textit{0.301} \\
      Margin@6 & \textit{0.242} \\
      Margin@7 & \textit{0.263}\\
      Margin@8 & \textit{0.256} \\
      Margin@9 & \textit{0.242} \\
      Margin@10 & \textit{0.236} \\
      Sequence Margin & \textit{0.281}\\
         \bottomrule
    \end{tabular}
    \caption{Pearson Correlation Coefficient between F1 scores and model uncertainty for \mistralthreesin. Statistics with $p<0.001$ are in \textit{italics}. The value in \textbf{bold} indicates the best result. Note, that negative correlation on the least confidence strategy is expected, since it represents model confidence rather that uncertainty.}
    \label{tab:pearson_corr}
\end{table}

\onecolumn
\section{Error Types Examples} \label{app:error-examples}
\begin{table*}[ht!]
    \centering
    \resizebox{\textwidth}{!}{%
    \begin{tabular}{p{0.13\linewidth}p{0.3\linewidth}p{0.4\linewidth}p{0.4\linewidth}p{0.4\linewidth}}
         \toprule
         \textbf{Error \mbox{Type}} & \textbf{User Query} & \textbf{Standing \mbox{Instructions}} & \textbf{Target} & \textbf{Prediction} \\
         \midrule
         
         Semantic Substitution & User: I want to find an apartment in Hayward. & > Request a home with one bed. & \texttt{GetHomes(area="Hayward", number\_of\_beds="1")} & \texttt{GetHomes({\color{red}{city}}="Hayward", number\_of\_beds=1)} \\

         \midrule
         Missing \mbox{Argument} & User: I am looking for an \colorbox{gray!20}{Gynecologist} in San Jose. & > Name Anjali Tate, M.D. as my preferred doctor when requesting a doctor. & \texttt{GetDoctors(\mbox{city="San Jose"}, \mbox{doctor\_name="Anjali Tate, M.D.",} {\color{blue}type="Gynecologist"})} & \texttt{GetDoctors(\mbox{city="San Jose"}, \mbox{doctor\_name="Anjali Tate,} M.D.")}\\

         \midrule
         Hallucination (new func.) & User: Can you let me know \colorbox{gray!20}{some attractions} to visit? & > If I'm looking to travel, my go-to spot is Chicago. \newline> If I'm looking into Travel, I should also check out Hotels. \newline> Request Hotels with a two-star rating. & \texttt{GetTravel(location="CHi-town")} \newline\texttt{GetHotels(average\_rating="2", location="CHi-town")} & \texttt{{\color{red} GetAttractions}(city="Chicago")} \newline \texttt{GetHotels(rating="2")}\\

         \midrule
         Hallucination (mixed calls) & User: Can you show some attractions to visit? \newline Agent: Sure. Where should I search for attractions in? \newline User: Find me something in Sydney, NSW please. & > I prefer the Museum category when requesting Travel. \newline > \colorbox{gray!20}{Choose a museum if you wish to have a} \colorbox{gray!20}{good experience with children.} \newline > I would like to request Travel for my preferred category of Park. & \texttt{GetTravel(\mbox{location="Sydney, NSW",} category="Museum", good\_for\_kids="True")} \newline \texttt{GetTravel(\mbox{location="Sydney, NSW",} category="Park")} & \texttt{GetTravel(\mbox{city="Sydney, NSW",} category="Museum")} \newline \texttt{GetTravel(\mbox{city="Sydney, NSW"}, category="Park", {\color{red}good\_for\_kids=True})}\\

         \midrule
         Combined Calls & User: I'm looking for Music events. & > If I'm looking for events, I'd like to check out what's going on in Portland. \newline > If I ask for Events, my preferred category is \colorbox{gray!20}{Blues or basketball}. & \texttt{GetEvents(category="Blues", city="Portland", event\_type="Music")} \newline \texttt{GetEvents(category="basketball", city="Portland", event\_type="Music")} &  \texttt{\color{red} GetEvents(city="Portland", genre=["Blues", "Basketball"])} \\

         \midrule
         Split Call & User: I want to find a new restaurant. What do you suggest to eat in San Francisco? & > Request Restaurants that serve Oriental cuisine. \newline > Search for the 8 Immortals Restaurant when looking for an Oriental restaurant. & \texttt{GetRestaurants(city="San \mbox{Francisco", cuisine="Oriental",} \mbox{restaurant\_name="8 Immortals} Restaurant")} & \texttt{\color{red} GetRestaurants(city="San Francisco", cuisine="Oriental")} \newline \texttt{\color{red}GetRestaurants(city="San \mbox{Francisco", restaurant\_name="8} Immortals")}\\

         \midrule 
         Wrong Value & User: Can you help me find some movies to watch online? & > Request \colorbox{gray!20}{funny} Media. & \texttt{GetMedia(genre="funny")} & \texttt{GetMedia(genre="{\color{red}Comedy}")}\\

         \midrule
         Value \mbox{Formatting} & User: I would like to rent a car from \colorbox{gray!20}{March 8th} in Paris, France. \newline Agent: At what time would you need it? And when is your return date? \newline User: I would need it at \colorbox{gray!20}{12 o'clock in the afternoon till} \colorbox{gray!20}{the 9th of this month}. & & \texttt{GetRentalCars(dropoff\_date="9th \mbox{of this month",} \mbox{pickup\_date="March 8th",} pickup\_time="12 o'clock", \newline ...} & \texttt{GetRentalCars(pickup\_time="{\color{red}12:00}", pickup\_date="{\color{red}2023-03-08}", dropoff\_date="{\color{red}2023-03-09}", \newline ...}\\

         \midrule
         Missing call & User: I need to find a General Practitioner doctor in San Jose. & > Request \colorbox{gray!20}{Access Health} as your doctor. \newline > If I ask for Doctor, my preferred doctor name is \colorbox{gray!20}{Daisy Manuel-Arguelles, DO}. & \texttt{\color{blue}GetDoctors(\mbox{city="San Jose",} \mbox{doctor\_name="Access Health",} type="General Practitioner")}\newline \texttt{GetDoctors(\mbox{city="San Jose",} \mbox{doctor\_name="Daisy Manuel}- \mbox{Arguelles, DO", type="General} Practitioner")} & \texttt{GetDoctors(\mbox{city="San Jose",} \mbox{type="General Practitioner",}  \mbox{doctor\_name="Daisy Manuel-} Arguelles, DO")}.  \\

         \midrule
         Dataset Error & User: I'm trying to find things to do. I'd like something in New York City. I like Electronica events and I'm looking for a Concert. \newline Agent: I found 3 events for you. One event is Crooked Colours at Rough Trade NYC. \newline User: Sure, that works for me. \colorbox{gray!20}{I'd like to find a room in a hotel} \colorbox{gray!20}{there.} & & \texttt{GetEvents(category="Electronica", \mbox{city="New York City",} \mbox{event\_name="Crooked Colours",} event\_type="Music")} & \texttt{GetEvents(\mbox{city="New York City",} event\_type="Concert", genre="Electronica")} \newline \texttt{\color{blue}GetHotels(\mbox{city="New York City",} location="Rough Trade NYC")} \\
        
         \bottomrule
    \end{tabular}}
    \caption{Examples of most prominent errors made by Mistral 3. Incorrectly predicted functions, arguments and values are marked in {\color{red}}. Missing arguments and API calls are in {\color{blue} blue}. Relevant parts of the user query and standing instructions are \colorbox{gray!20}{highlighted}.}
    \label{tab:error_types}
\end{table*}

\newpage
\section{Additional Results}\label{app:extra-results}
\begin{figure}[h]
    \centering
    \includegraphics[width=0.7\linewidth]{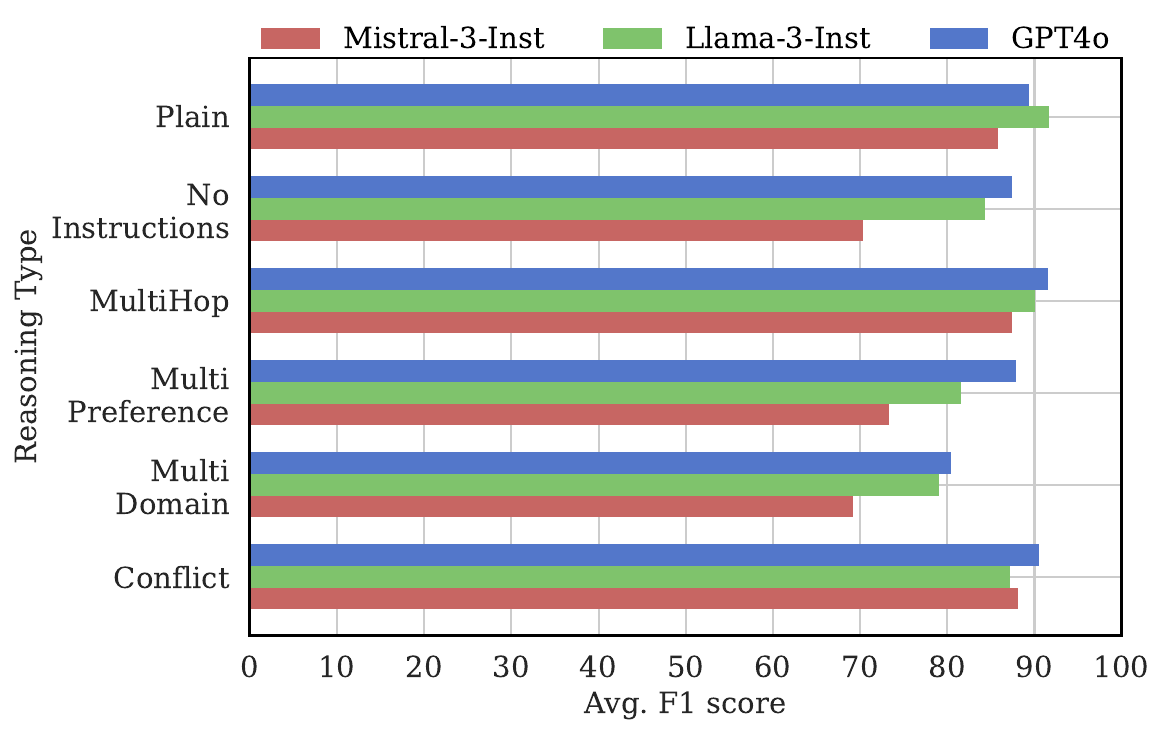}
    \caption{Average F1 scores of \name models per each reasoning type.}
    \label{fig:example_types_final}
\end{figure}

\begin{figure}[h]
    \centering
    \includegraphics[width=0.7\linewidth]{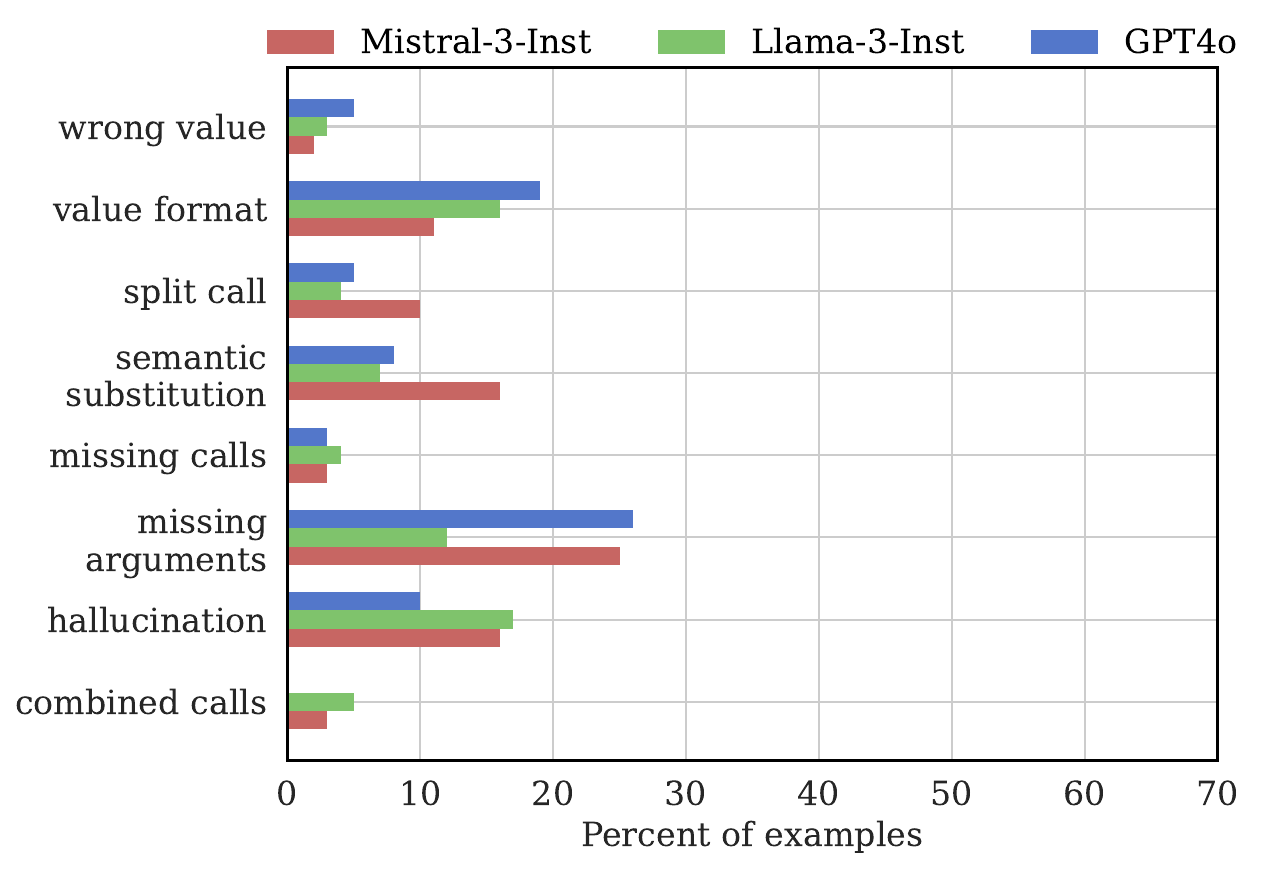}
    \caption{Distribution of errors on a sample of \name's
predictions.}
    \label{fig:error_types_final}
\end{figure}

\begin{figure*}[th!]
    \centering
    \subfloat[EM]{\includegraphics[width=0.32\textwidth]{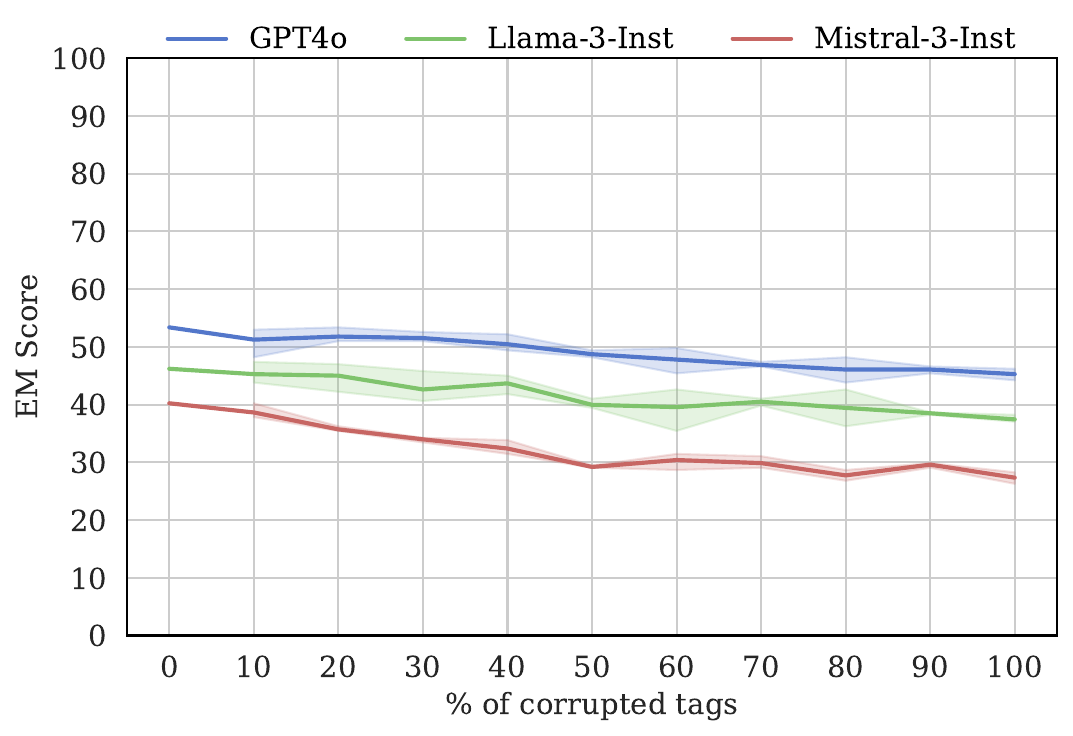}}
    \subfloat[Precision]{\includegraphics[width=0.32\textwidth]{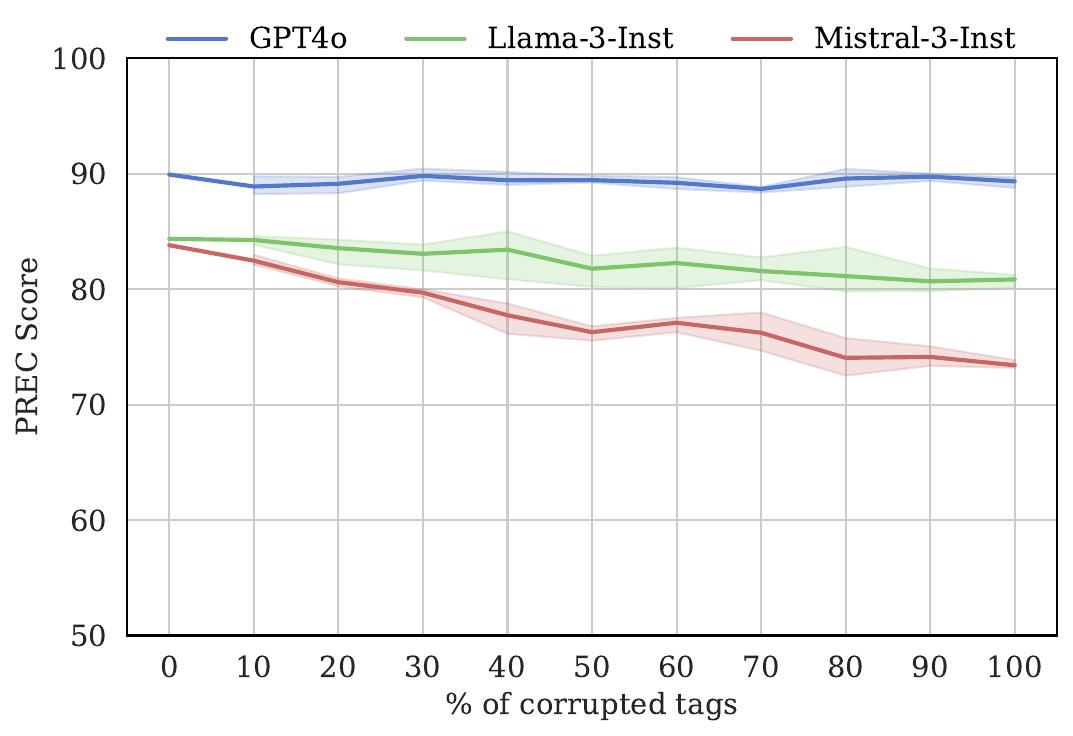}}
    \subfloat[Recall]{\includegraphics[width=0.32\textwidth]{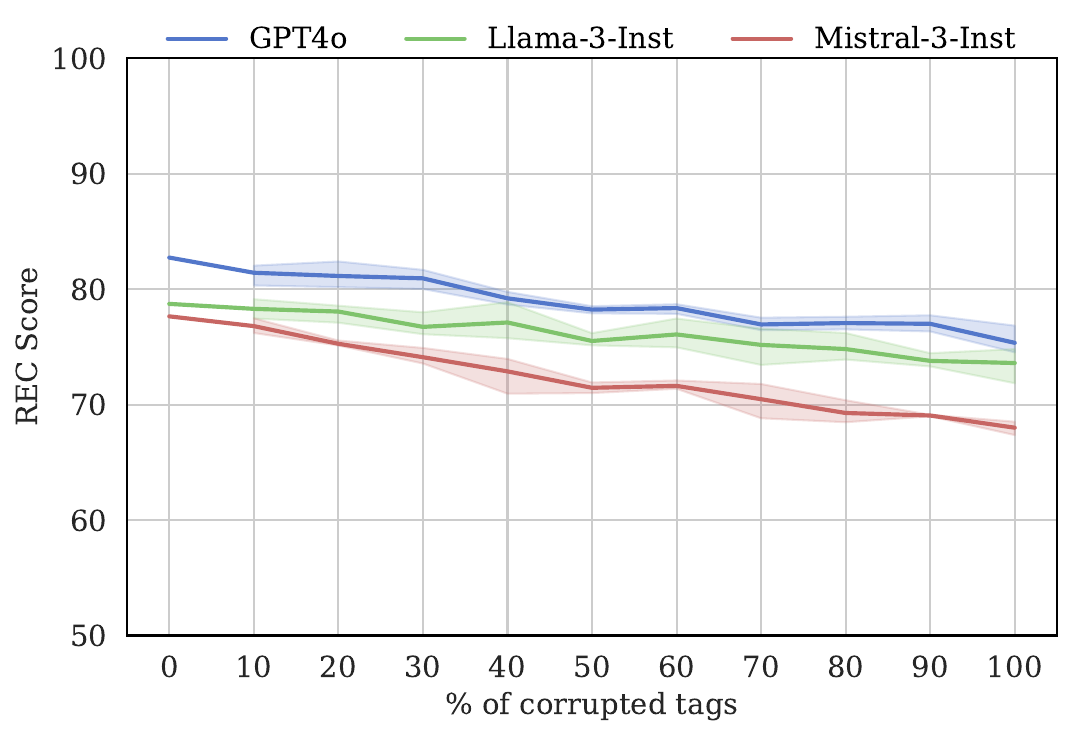}}
    \caption{The effects of tag quality on the downstream task scores of \gptb, \llamathreeschat, and \mistralthreesin. All scores are in \%.}
    \label{fig:tag_effects}
\end{figure*}

\newpage
\section{Prompts} \label{app:prompts}
All of the prompts we use follow the same structure: \colorbox{gray!20}{Task Description} + \colorbox{yellow!30}{API Schema} + \colorbox{blue!15}{Input Description} (optionally) + \colorbox{red!10}{Example(s)}. We provide the list of prompts below.

\subsection{Baseline Prompt} \label{app:baseline-prompt}

\noindent System prompt template:
\begin{tcolorbox}[colback=gray!20,colframe=gray!20,boxsep=-1mm,breakable]
    \scriptsize
    You are designing a parser that takes in a user utterance and some standing instructions and outputs a set of API calls. 
    \\Every API call consists of "GetX" where X is domain name and uses slot names listed below as arguments.  We list the domain name followed by the list of possible slot names. Some slot names can be categorical or boolean
    \\The values for the arguments can come from the user's dialogue or standing instructions. If the user requests a slot name and no value is found, use "?". If the user requests dontcare, use value as "any".
    \\Standing instructions allow you to add preferences or requirements that you’d like to consider when generating the parser. 
    \\If standing instructions are applicable across multiple domains, place an API call per situation per domain. 
    \\If some of the applicable standing instructions have instructions of similar type, place multiple API calls respecting the standing instructions.
    \\If some slots are applicable across several domains, generate the respective slot names for the respective domains.
\end{tcolorbox}
\vspace{-1em}
\begin{tcolorbox}[colback=yellow!30,colframe=yellow!30,boxsep=-1mm,breakable]
    \scriptsize
    Schema:
    \\Banks: recipient\_account\_name, amount, recipient\_account\_type
    \\ Buses: origin, departure\_date, fare\_type, transfers, price, group\_size, destination, destination\_station\_name, origin\_station\_name, departure\_time
    \\ Events: event\_name, city, category, event\_location, number\_of\_tickets, time, address\_of\_location, date, venue\_address, event\_type
    \\Flights: origin, inbound\_arrival\_time, is\_redeye, outbound\_departure\_time, outbound\_arrival\_time, inbound\_departure\_time, return\_date, airlines, seating\_class, refundable, number\_stops, destination\_airport, departure\_date, fare, destination, passengers, origin\_airport
    \\Homes: pets\_allowed, visit\_date, address, property\_name, rent, number\_of\_baths, area, number\_of\_beds, furnished, phone\_number
    \\Hotels: has\_wifi, average\_rating, check\_out\_date, price, pets\_welcome, number\_of\_days, location, check\_in\_date, phone\_number, number\_of\_rooms, street\_address, hotel\_name
    \\HouseStays: rating, phone\_number, has\_laundry\_service, check\_out\_date, total\_price, check\_in\_date, address, number\_of\_adults, where\_to
    \\Media: title, directed\_by, subtitles, genre
    \\Movies: theater\_name, movie\_name, price, show\_date, location, show\_time, number\_of\_tickets, genre, show\_type, street\_address
    \\Music: song\_name, year, album, artist, genre, playback\_device
    \\RentalCars: dropoff\_date, pickup\_time, pickup\_city, pickup\_date, total\_price, car\_type, car\_name, pickup\_location
    \\Restaurants: price\_range, restaurant\_name, city, has\_live\_music, serves\_alcohol, time, date, phone\_number, cuisine, street\_address, party\_size
    \\Salons: is\_unisex, average\_rating, city, appointment\_date, appointment\_time, stylist\_name, phone\_number, street\_address
    \\Dentists: dentist\_name, phone\_number, offers\_cosmetic\_services, city, appointment\_date, appointment\_time, address
    \\Doctors: doctor\_name, city, average\_rating, appointment\_date, appointment\_time, type, phone\_number, street\_address
    \\Travel: good\_for\_kids, category, attraction\_name, location, phone\_number, free\_entry
    \\Weather: city, temperature, date, precipitation, humidity, wind

    \hfill \\
    Further, following slots have categorical values:
    \\recipient\_account\_type: checking, savings
    \\fare\_type: Economy, Economy extra, Flexible
    \\(Travel) category: Place of Worship, Theme Park, Museum, Historical Landmark, Park, Tourist Attraction, Sports Venue, Shopping Area, Performing Arts Venue, Nature Preserve
    \\event\_type: Music, Sports
    \\seating\_class: Economy, Premium Economy, Business, First Class
    \\refundable: True, False
    \\airlines: United Airlines, American Airlines, Delta Airlines, Southwest Airlines, Alaska Airlines, British Airways, Air Canada, Air France
    \\show\_type: regular, 3d, imax
    \\playback\_device: TV, kitchen speaker, bedroom speaker
    \\(Doctors) type: Gynecologist, ENT Specialist, Ophthalmologist, General Practitioner, Dermatologist
    \\car\_type: Compact, Standard, Full-size
    \\price\_range: inexpensive, moderate, expensive, very expensive

    \hfill \\
    Further, following slots are boolean:
    \\has\_wifi, pets\_allowed, subtitles, offers\_cosmetic\_services, has\_laundry\_service, is\_unisex, good\_for\_kids, has\_live\_music, pets\_welcome, serves\_alcohol, is\_redeye, furnished, free\_entry
\end{tcolorbox}

\noindent Example template: 
\begin{tcolorbox}[colback=red!10,colframe=red!10,boxsep=-1mm,breakable]
    \scriptsize
    Dialogue:
    \\ \texttt{{\color{blue}\{\{} user\_utterance {\color{blue}\}\}}}\\
    \hfill \\
    Applicable Standing Instructions:
    \\ \texttt{{\color{blue}\{\{} applicable\_instructions | join({\color{red}{\dbq}$\backslash$n> {\dbq}}) {\color{blue}\}\}}} \\
    \hfill \\
    API Calls: 
\end{tcolorbox}
\noindent Target template:
\begin{tcolorbox}[colback=red!10,colframe=red!10,boxsep=-1mm, breakable]
    \scriptsize
    \texttt{{\color{blue}\{\{} target\_api\_calls | join({\color{red}{\dbq}$\backslash$n{\dbq}}) {\color{blue}\}\}}}
\end{tcolorbox}

\subsection{\textsc{Default} Prompt V1} \label{app:def-prompt}

\noindent System prompt template:
\begin{tcolorbox}[colback=gray!20,colframe=gray!20,boxsep=-1mm,breakable]
    \scriptsize
    You are designing a parser that takes in a user utterance and some standing instructions and outputs a set of API calls. 
    \\Every API call consists of "GetX" where X is domain name and uses slot names listed below as arguments.  We list the domain name followed by the list of possible slot names. Some slot names can be categorical or boolean
    \\The values for the arguments can come from the user's dialogue or standing instructions. If the user requests a slot name and no value is found, use "?". If the user requests dontcare, use value as "any".
    \\Standing instructions allow you to add preferences or requirements that you’d like to consider when generating the parser. 
    \\If standing instructions are applicable across multiple domains, place an API call per situation per domain. 
    \\If some of the applicable standing instructions have instructions of similar type, place multiple API calls respecting the standing instructions.
    \\If some slots are applicable across several domains, generate the respective slot names for the respective domains.
\end{tcolorbox}
\vspace{-1em}
\begin{tcolorbox}[colback=yellow!30,colframe=yellow!30,boxsep=-1mm,breakable]
    \scriptsize
    Schema:
    \\Banks: recipient\_account\_name, amount, recipient\_account\_type
    \\ Buses: origin, departure\_date, fare\_type, transfers, price, group\_size, destination, destination\_station\_name, origin\_station\_name, departure\_time
    \\ Events: event\_name, city, category, event\_location, number\_of\_tickets, time, address\_of\_location, date, venue\_address, event\_type
    \\Flights: origin, inbound\_arrival\_time, is\_redeye, outbound\_departure\_time, outbound\_arrival\_time, inbound\_departure\_time, return\_date, airlines, seating\_class, refundable, number\_stops, destination\_airport, departure\_date, fare, destination, passengers, origin\_airport
    \\Homes: pets\_allowed, visit\_date, address, property\_name, rent, number\_of\_baths, area, number\_of\_beds, furnished, phone\_number
    \\Hotels: has\_wifi, average\_rating, check\_out\_date, price, pets\_welcome, number\_of\_days, location, check\_in\_date, phone\_number, number\_of\_rooms, street\_address, hotel\_name
    \\HouseStays: rating, phone\_number, has\_laundry\_service, check\_out\_date, total\_price, check\_in\_date, address, number\_of\_adults, where\_to
    \\Media: title, directed\_by, subtitles, genre
    \\Movies: theater\_name, movie\_name, price, show\_date, location, show\_time, number\_of\_tickets, genre, show\_type, street\_address
    \\Music: song\_name, year, album, artist, genre, playback\_device
    \\RentalCars: dropoff\_date, pickup\_time, pickup\_city, pickup\_date, total\_price, car\_type, car\_name, pickup\_location
    \\Restaurants: price\_range, restaurant\_name, city, has\_live\_music, serves\_alcohol, time, date, phone\_number, cuisine, street\_address, party\_size
    \\Salons: is\_unisex, average\_rating, city, appointment\_date, appointment\_time, stylist\_name, phone\_number, street\_address
    \\Dentists: dentist\_name, phone\_number, offers\_cosmetic\_services, city, appointment\_date, appointment\_time, address
    \\Doctors: doctor\_name, city, average\_rating, appointment\_date, appointment\_time, type, phone\_number, street\_address
    \\Travel: good\_for\_kids, category, attraction\_name, location, phone\_number, free\_entry
    \\Weather: city, temperature, date, precipitation, humidity, wind

    \hfill \\
    Further, following slots have categorical values:
    \\recipient\_account\_type: checking, savings
    \\fare\_type: Economy, Economy extra, Flexible
    \\(Travel) category: Place of Worship, Theme Park, Museum, Historical Landmark, Park, Tourist Attraction, Sports Venue, Shopping Area, Performing Arts Venue, Nature Preserve
    \\event\_type: Music, Sports
    \\seating\_class: Economy, Premium Economy, Business, First Class
    \\refundable: True, False
    \\airlines: United Airlines, American Airlines, Delta Airlines, Southwest Airlines, Alaska Airlines, British Airways, Air Canada, Air France
    \\show\_type: regular, 3d, imax
    \\playback\_device: TV, kitchen speaker, bedroom speaker
    \\(Doctors) type: Gynecologist, ENT Specialist, Ophthalmologist, General Practitioner, Dermatologist
    \\car\_type: Compact, Standard, Full-size
    \\price\_range: inexpensive, moderate, expensive, very expensive

    \hfill \\
    Further, following slots are boolean:
    \\has\_wifi, pets\_allowed, subtitles, offers\_cosmetic\_services, has\_laundry\_service, is\_unisex, good\_for\_kids, has\_live\_music, pets\_welcome, serves\_alcohol, is\_redeye, furnished, free\_entry
\end{tcolorbox}

\vspace{-1em}

\begin{tcolorbox}[colback=blue!15,colframe=blue!15,boxsep=-1mm,breakable]
    \scriptsize
    ---\\
    \hfill \\
    \texttt{{\color{blue}\{\% if} model\_name == {\color{red}{\dbq}llama\dbq\xspace}{\color{blue}\%\}}}\\
    Follow the following format.\\
    \texttt{\color{blue}\{\% else \%\}}\\
    The examples are formatted as follows.\\
    \texttt{\color{blue}\{\% endif \%\}}\\
    
    Dialogue:\\
    <user\_utterance>\\
    
    Applicable Standing Instructions:\\
    <applicable\_standing\_instructions>\\ 
    
    API Calls: \\
    API calls to solve the user task\\

    ---\\

    \texttt{{\color{blue}\{\% if} model\_name == {\color{red}{\dbq}llama\dbq\xspace}{\color{blue}\%\}}}\\
    You are given several independent examples of the task:\\
    \texttt{\color{blue}\{\% endif \%\}}\\
    
\end{tcolorbox}

\noindent Example template: 
\begin{tcolorbox}[colback=red!10,colframe=red!10,boxsep=-1mm,breakable]
    \scriptsize
    \texttt{{\color{blue}\{\% if} split == {\color{red}{\dbq}test\dbq\xspace} {\color{blue}and} model\_name == {\color{red}{\dbq}llama\dbq\xspace}{\color{blue}\%\}}}\\
    Given the examples above, output only the API calls for the following example with no additional text:\\
    \texttt{\color{blue}\{\% endif \%\}}\\
    
    Dialogue:
    \\ \texttt{{\color{blue}\{\{} user\_utterance {\color{blue}\}\}}}\\
    \hfill \\
    Applicable Standing Instructions:
    \\ \texttt{{\color{blue}\{\{} applicable\_instructions | join({\color{red}{\dbq}$\backslash$n> {\dbq}}) {\color{blue}\}\}}} \\
    \hfill \\
    API Calls: 
\end{tcolorbox}
\noindent Target template:
\begin{tcolorbox}[colback=red!10,colframe=red!10,boxsep=-1mm]
    \scriptsize
    \texttt{{\color{blue}\{\{} target\_api\_calls | join({\color{red}{\dbq}$\backslash$n{\dbq}}) {\color{blue}\}\}}}
\end{tcolorbox}

\subsection{\textsc{Default} Prompt V2} \label{app:def-prompt-2}

\noindent System prompt template:

\begin{tcolorbox}[colback=gray!20,colframe=gray!20,boxsep=-1mm]
    \scriptsize
    \textnormal{You are designing a parser that takes in a user utterance (field `user\_utterance`) and a user profile with standing instructions (field `user\_profile`) and outputs a set of API calls as an answer.}\\
\textnormal{Every API call consist of "GetX" where X is domain name and uses slot names listed below as arguments.  We list the domain name followed by the list of possible slot names in the `api\_schema` field. Some slot names can be categorical or boolean.}\\
\textnormal{The values for the arguments can come from the user's dialogue or standing instructions. If the user asks about a slot but no value is found, set its value to "?". If the user explicitly says they do not care about a particular slot, set its value to "any".}\\
Standing instructions allow you to add preferences or requirements that you'd like to consider when generating the parser.\\
If standing instructions are applicable across multiple domains, place an API call per situation per domain.\\
If some of the applicable standing instructions have instructions of similar type, place multiple API calls respecting the standing instructions.\\
If some slots are applicable across several domains, generate the respective slot names for the respective domains.\\

\end{tcolorbox}
\noindent The schema template, input description and example formatting are the same as in \autoref{app:def-prompt-2}

\subsection{\textsc{External Tag} (\extag) Prompt V1} \label{app:tag-simple-prompt}
\noindent System prompt template:

\begin{tcolorbox}[colback=gray!20,colframe=gray!20,boxsep=-1mm,breakable]
    \scriptsize
    You are designing a parser that takes in a user query and some user preferences and outputs a set of API calls. Execution of the API calls helps to answer the user query.
\\Every function name in the API call has a structure of "GetX" where X is domain name. Each function uses slot names listed below as arguments. Some slot names can be categorical or boolean. The values for the arguments can come from the user's query or user preferences. If the user requests a slot name and no value is found, use "?". If the user says they don't care, set slot value to "any".
\\User preferences allow you to add preferences or requirements that you’d like to consider when generating the parser. If user preferences are applicable across multiple domains, place an API call per situation per domain. If some of the applicable preferences have instructions of similar type, place multiple API calls respecting the preferences. If some slots are applicable across several domains, generate the respective slot names for the respective domains.\\

The user profile would be tagged in the following format: \\
    \text{<a:API\_FUNCTION\_NAME> text </a>} would mean the function that is relevant for the text in brackets \\
    \text{<sl:SLOT\_NAME> value </sl>} would highlight which function arguments are used in the function and their value.

    \hfill \\
    Output a list of API calls that would answer the user query. There can be several API calls per user query, but not always, so output only the required calls. Make sure you follow the following output structure: GetX(slot1="value1", slot2="value2"). Use the tags from the user profile, as well as information from the current dialogue to generate the calls. In cases, where seceral API calls are required, generate each one in a new line. Use only the functions from the documentation above, and make sure to check that only the slots for the chosen function are used in the API call. Generate only the API calls.
\end{tcolorbox}
\vspace{-1em}
\begin{tcolorbox}[colback=yellow!30,colframe=yellow!30,boxsep=-1mm,breakable]
    \scriptsize
    The list of the available function names is presented below, followed by possible slot names.
    \\Some of the possible API calls include functions:
    \\GetBanks: handling all the banking information (recipient\_account\_name, amount, recipient\_account\_type)
    \\ GetBuses: finding and booking bus tickets and routes (origin, departure\_date, fare\_type, transfers, price, group\_size, destination, departure\_time)
    \\ GetEvents: finding and booking events (event\_name, city, category, number\_of\_tickets, time, date, venue\_address, event\_type)
    \\ GetFlights: finding and booking flights (origin, inbound\_arrival\_time, is\_redeye, outbound\_departure\_time, outbound\_arrival\_time, inbound\_departure\_time, return\_date, airlines, seating\_class, refundable, number\_stops, departure\_date, fare, destination, passengers)
    \\ GetHomes: looking for property (pets\_allowed, visit\_date, address, property\_name, rent, number\_of\_baths, area, number\_of\_beds, furnished, phone\_number)
    \\ GetHotels: booking hotels (has\_wifi, average\_rating, check\_out\_date, price, pets\_welcome, number\_of\_days, location, check\_in\_date, phone\_number, number\_of\_rooms, street\_address, hotel\_name)
    \\ GetHouseStays: booking temporary accommodation (rating, phone\_number, has\_laundry\_service, check\_out\_date, total\_price, check\_in\_date, address, number\_of\_adults, where\_to)
    \\ GetMedia: searching for online media (title, directed\_by, subtitles, genre)
    \\ GetMovies: searching for cinema tickets (theater\_name, movie\_name, price, show\_date, location, show\_time, number\_of\_tickets, genre, show\_type, street\_address)
    \\ GetMusic: finding songs (song\_name, year, album, artist, genre, playback\_device)
    \\ GetRentalCars: booking rental cars (dropoff\_date, pickup\_time, pickup\_city, pickup\_date, total\_price, car\_type, car\_name, pickup\_location)
    \\GetRestaurants: finding and booking restaurants (price\_range, restaurant\_name, city, has\_live\_music, serves\_alcohol, time, date, phone\_number, cuisine, street\_address, party\_size)
    \\GetSalons: finding hair salons (is\_unisex, average\_rating, city, appointment\_date, appointment\_time, stylist\_name, phone\_number, street\_address)
    \\GetDentists: finding dentists (dentist\_name, phone\_number, offers\_cosmetic\_services, city, appointment\_date, appointment\_time, address)
    \\GetDoctors: finding doctors (doctor\_name, city, average\_rating, appointment\_date, appointment\_time, type, phone\_number, street\_address)
    \\GetTravel: finding attractions (good\_for\_kids, category, attraction\_name, location, phone\_number, free\_entry)
    \\GetWeather: getting weather information (city, temperature, date, precipitation, humidity, wind)

    \hfill \\
    Further, following slots have categorical values:
    \\recipient\_account\_type: checking, savings
    \\fare\_type: Economy, Economy extra, Flexible
    \\(Travel) category: Place of Worship, Theme Park, Museum, Historical Landmark, Park, Tourist Attraction, Sports Venue, Shopping Area, Performing Arts Venue, Nature Preserve
    \\event\_type: Music, Sports
    \\seating\_class: Economy, Premium Economy, Business, First Class
    \\refundable: True, False
    \\airlines: United Airlines, American Airlines, Delta Airlines, Southwest Airlines, Alaska Airlines, British Airways, Air Canada, Air France
    \\show\_type: regular, 3d, imax
    \\playback\_device: TV, kitchen speaker, bedroom speaker
    \\(Doctors) type: Gynecologist, ENT Specialist, Ophthalmologist, General Practitioner, Dermatologist
    \\car\_type: Compact, Standard, Full-size
    \\price\_range: inexpensive, moderate, expensive, very expensive

    \hfill \\
    Further, following slots are boolean:
    \\has\_wifi, pets\_allowed, subtitles, offers\_cosmetic\_services, has\_laundry\_service, is\_unisex, good\_for\_kids, has\_live\_music, pets\_welcome, serves\_alcohol, is\_redeye, furnished, free\_entry

\end{tcolorbox}
\vspace{-1em}
\begin{tcolorbox}[colback=blue!15,colframe=blue!15,boxsep=-1mm,breakable]
    \scriptsize
    ---\\
    \hfill \\
    \texttt{{\color{blue}\{\% if} model\_name == {\color{red}{\dbq}llama\dbq\xspace}{\color{blue}\%\}}}\\
    Follow the following format.\\
    \texttt{\color{blue}\{\% else \%\}}\\
    The examples are formatted as follows.\\
    \texttt{\color{blue}\{\% endif \%\}}\\
    
    Dialogue:\\
    <user\_utterance>\\
    
    Applicable Standing Instructions:\\
    <applicable\_standing\_instructions>\\ 

    Tagged Standing Instructions:\\
    <tagged applicable standing instructions>\\
    
    API Calls: \\
    API calls to solve the user task\\

    ---\\

    \texttt{{\color{blue}\{\% if} model\_name == {\color{red}{\dbq}llama\dbq\xspace}{\color{blue}\%\}}}\\
    You are given several independent examples of the task:\\
    \texttt{\color{blue}\{\% endif \%\}}
\end{tcolorbox}

\noindent Example template: 

\begin{tcolorbox}[colback=red!10,colframe=red!10,boxsep=-1mm,breakable]
    \scriptsize
    \texttt{{\color{blue}\{\% if} split == {\color{red}{\dbq}test\dbq\xspace} {\color{blue}and} model\_name == {\color{red}{\dbq}llama\dbq\xspace}{\color{blue}\%\}}}\\
    Given the examples above, output only the API calls for the following example with no additional text:\\
    \texttt{\color{blue}\{\% endif \%\}}\\
    
    Dialogue:
    \\ \texttt{{\color{blue}\{\{} user\_utterance {\color{blue}\}\}}}\\
    \hfill \\
    Applicable Standing Instructions:
    \\ \texttt{{\color{blue}\{\{} applicable\_instructions | join({\color{red}{\dbq}$\backslash$n> {\dbq}}) {\color{blue}\}\}}} \\
    \hfill \\
    Tagged Applicable Standing Instructions:
    \\ \texttt{{\color{blue}\{\{} tagged\_applicable\_instructions | join({\color{red}{\dbq}$\backslash$n> {\dbq}}) {\color{blue}\}\}}} \\
    \hfill \\
    API Calls: 
\end{tcolorbox}

\noindent Target template:

\begin{tcolorbox}[colback=red!10,colframe=red!10,boxsep=-1mm,breakable]
    \scriptsize
    \texttt{{\color{blue}\{\{} target\_api\_calls | join({\color{red}{\dbq}$\backslash$n{\dbq}}) {\color{blue}\}\}}}
\end{tcolorbox}

\vspace{-1em}
\subsection{\textsc{External Tag} (\extag) Prompt V2} \label{app:tag-simple-prompt-2}

\noindent System prompt template:

\begin{tcolorbox}[colback=gray!20,colframe=gray!20,boxsep=-1mm,breakable]
    \scriptsize
    You are a parser that converts user queries and profile preferences into API calls to fulfill the query. Use the provided tags, dialogue, and schema to generate precise API calls.  \\

**Task Guidelines:**  \\
1. **API Call Structure:**\\  
   \hspace*{1em}\textnormal{Format each call as `GetX(slot1="value1", slot2="value2", ...)`, where `X` is the domain name, and slots match the chosen function.}  \\

2. **Using Tags:**  \\
   \hspace*{1em}\textnormal{- `<a:API\_FUNCTION\_NAME>` marks relevant functions.}\\
   \hspace*{1em}\textnormal{- `<sl:SLOT\_NAME>` specifies slot values.}\\ 
   \hspace*{1em}\textnormal{Example: `<a:GET\_FLIGHTS> Request <sl:AIRLINES> Alaska Airlines</sl></a>` becomes `airlines="Alaska Airlines"`.}  \\

3. **Slot Values:**  \\
   \hspace*{1em}- Use values from the query or tags.\\  
   \hspace*{1em}\textnormal{- Assign `"?"` if a slot is missing and `"any"` if the user has no preference. }\\

4. **Output Requirements:**  \\
   \hspace*{1em}- Include only required API calls.  \\
   \hspace*{1em}- Output each call on a new line.  \\

---

\end{tcolorbox}
\vspace{-1em}
\begin{tcolorbox}[colback=yellow!30,colframe=yellow!30,boxsep=-1mm,breakable]
    \scriptsize
    **Schema:** \\ 
Use valid functions and slots as listed:  \\

\#\#\#\# **Functions and Slots**  \\
Each function corresponds to a specific domain and has associated slots. Use only the listed slots for each function.  \\

- **GetBanks**  \\
  \hspace*{1em}\textnormal{- Slots: `recipient\_account\_name`, `amount`, `recipient\_account\_type`} \\

- **GetBuses**  \\
  \hspace*{1em}\textnormal{- Slots: `origin`, `departure\_date`, `fare\_type`, `transfers`, `price`, `group\_size`, `destination`, `departure\_time`}  \\

- **GetEvents**  \\
  \hspace*{1em}\textnormal{- Slots: `event\_name`, `city`, `category`, `number\_of\_tickets`, `time`, `date`, `venue\_address`, `event\_type`}  \\

- **GetFlights**  \\
  \hspace*{1em}\textnormal{- Slots: `origin`, `inbound\_arrival\_time`, `is\_redeye`, `outbound\_departure\_time`, `outbound\_arrival\_time`, `inbound\_departure\_time`, `return\_date`, `airlines`, `seating\_class`, `refundable`, `number\_stops`, `departure\_date`, `fare`, `destination`, `passengers`}\\  

- **GetHomes**  \\
  \hspace*{1em}\textnormal{- Slots: `pets\_allowed`, `visit\_date`, `address`, `property\_name`, `rent`, `number\_of\_baths`, `area`, `number\_of\_beds`, `furnished`, `phone\_number`}\\  

- **GetHotels**  \\
  \hspace*{1em}\textnormal{- Slots: `has\_wifi`, `average\_rating`, `check\_out\_date`, `price`, `pets\_welcome`, `number\_of\_days`, `location`, `check\_in\_date`, `phone\_number`, `number\_of\_rooms`, `street\_address`, `hotel\_name`}  \\

- **GetHouseStays**  \hfill \\
  \hspace*{1em}\textnormal{- Slots: `rating`, `phone\_number`, `has\_laundry\_service`, `check\_out\_date`, `total\_price`, `check\_in\_date`, `address`, `number\_of\_adults`, `where\_to`}  \\
    \hfill 
- **GetMedia**  \\
  \hspace*{1em}\textnormal{- Slots: `title`, `directed\_by`, `subtitles`, `genre`} \\ 
    
- **GetMovies**  \hfill \\
  \hspace*{1em}\textnormal{- Slots: `theater\_name`, `movie\_name`, `price`, `show\_date`, `location`, `show\_time`, `number\_of\_tickets`, `genre`, `show\_type`, `street\_address`} \\

- **GetMusic**  \\
  \hspace*{1em}\textnormal{- Slots: `song\_name`, `year`, `album`, `artist`, `genre`, `playback\_device`}  \\

- **GetRentalCars**  \\
  \hspace*{1em}\textnormal{- Slots: `dropoff\_date`, `pickup\_time`, `pickup\_city`, `pickup\_date`, `total\_price`, `car\_type`, `car\_name`, `pickup\_location`} \\ 

- **GetRestaurants**  \\
  \hspace*{1em}\textnormal{- Slots: `price\_range`, `restaurant\_name`, `city`, `has\_live\_music`, `serves\_alcohol`, `time`, `date`, `phone\_number`, `cuisine`, `street\_address`, `party\_size`}  \\

- **GetSalons**  \\
  \hspace*{1em}\textnormal{- Slots: `is\_unisex`, `average\_rating`, `city`, `appointment\_date`, `appointment\_time`, `stylist\_name`, `phone\_number`, `street\_address`} \\ 

- **GetDentists**  \\
  \hspace*{1em}\textnormal{- Slots: `dentist\_name`, `phone\_number`, `offers\_cosmetic\_services`, `city`, `appointment\_date`, `appointment\_time`, `address`}  \\

- **GetDoctors**  \\
  \hspace*{1em}\textnormal{- Slots: `doctor\_name`, `city`, `average\_rating`, `appointment\_date`, `appointment\_time`, `type`, `phone\_number`, `street\_address`}  \\

- **GetTravel** \\ 
  \hspace*{1em}\textnormal{- Slots: `good\_for\_kids`, `category`, `attraction\_name`, `location`, `phone\_number`, `free\_entry`}  \\

- **GetWeather**  \\
  \hspace*{1em}\textnormal{- Slots: `city`, `temperature`, `date`, `precipitation`, `humidity`, `wind`}  \\

---\\

\#\#\# **Slot Value Types**\\

\#\#\#\# **Categorical Slots**  \\
- `recipient\_account\_type`: `checking`, `savings` \\ 
- `fare\_type`: `Economy`, `Economy extra`, `Flexible`  \\
- `category` (Travel): `Place of Worship`, `Theme Park`, `Museum`, `Historical Landmark`, `Park`, `Tourist Attraction`, `Sports Venue`, `Shopping Area`, `Performing Arts Venue`, `Nature Preserve`  \\
- `event\_type`: `Music`, `Sports`  \\
- `seating\_class`: `Economy`, `Premium Economy`, `Business`, `First Class`  \\
- `refundable`: `True`, `False`  \\
- `airlines`: `United Airlines`, `American Airlines`, `Delta Airlines`, `Southwest Airlines`, `Alaska Airlines`, `British Airways`, `Air Canada`, `Air France`  \\
- `show\_type`: `regular`, `3d`, `imax`  \\
- `playback\_device`: `TV`, `kitchen speaker`, `bedroom speaker`  \\
- `type` (Doctors): `Gynecologist`, `ENT Specialist`, `Ophthalmologist`, `General Practitioner`, `Dermatologist`  \\
- `car\_type`: `Compact`, `Standard`, `Full-size`  \\
- `price\_range`: `inexpensive`, `moderate`, `expensive`, `very expensive`  \\

\#\#\#\# **Boolean Slots** \\ 
- `has\_wifi`, `pets\_allowed`, `subtitles`, `offers\_cosmetic\_services`, `has\_laundry\_service`, `is\_unisex`, `good\_for\_kids`, `has\_live\_music`, `pets\_welcome`, `serves\_alcohol`, `is\_redeye`, `furnished`, `free\_entry`  \\

---\\

Ensure all outputs strictly adhere to the required format and schema. Generate only API calls.

\end{tcolorbox}

\noindent The input description and example templates are the same as in \autoref{app:tag-simple-prompt}

\subsection{\jointtag\xspace Prompt} \label{app:tag-gen-prompt}

\noindent System prompt template:

\begin{tcolorbox}[colback=gray!20,colframe=gray!20,boxsep=-1mm,breakable]
    \scriptsize
    You are designing a parser that takes in a user utterance and some standing instructions and outputs a set of API calls. \\
\textnormal{Every API call consist of "GetX" where X is domain name and uses slot names listed below as arguments.  We list the domain name followed by the list of possible slot names. Some slot names can be categorical or boolean}\\
\textnormal{The values for the arguments can come from the user's dialogue or standing instructions. If the user asks about a slot but no value is found, set its value to "?". If the user explicitly says they do not care about a particular slot, set its value to "any".}\\
Standing instructions allow you to add preferences or requirements that you'd like to consider when generating the parser.\\
If standing instructions are applicable across multiple domains, place an API call per situation per domain.\\
If some of the applicable standing instructions have instructions of similar type, place multiple API calls respecting the standing instructions.\\
If some slots are applicable across several domains, generate the respective slot names for the respective domains.\\

Think step by step.\\
First, identify and label API calls and their slots within applicable standing instructions.\\
\textnormal{Use action tags such as <a:API\_NAME> ... </a>, and nested tags denoting specific attributes, like <sl:SLOT\_NAME> ... </sl>.}\\
Ensure that all tags are correctly placed, slot and API names are correct, all original sentence tokens are present and are in the correct order, no additional tokens are added, and slot values include only necessary information, e.g. the value of the slot.\\
Use those generated labels, as well as information from the dialogue to create the calls.\\
After that, output a list of API calls that would answer the user query.
\end{tcolorbox}
\vspace{-1em}
\begin{tcolorbox}[%
        enhanced, 
        colback=yellow!30,
        colframe=yellow!30,
        boxsep=-1mm,
        breakable
    ]
    \scriptsize
    Schema:
    \\Banks: recipient\_account\_name, amount, recipient\_account\_type
    \\ Buses: origin, departure\_date, fare\_type, transfers, price, group\_size, destination, destination\_station\_name, origin\_station\_name, departure\_time
    \\ Events: event\_name, city, category, event\_location, number\_of\_tickets, time, address\_of\_location, date, venue\_address, event\_type
    \\Flights: origin, inbound\_arrival\_time, is\_redeye, outbound\_departure\_time, outbound\_arrival\_time, inbound\_departure\_time, return\_date, airlines, seating\_class, refundable, number\_stops, destination\_airport, departure\_date, fare, destination, passengers, origin\_airport
    \\Homes: pets\_allowed, visit\_date, address, property\_name, rent, number\_of\_baths, area, number\_of\_beds, furnished, phone\_number
    \\Hotels: has\_wifi, average\_rating, check\_out\_date, price, pets\_welcome, number\_of\_days, location, check\_in\_date, phone\_number, number\_of\_rooms, street\_address, hotel\_name
    \\HouseStays: rating, phone\_number, has\_laundry\_service, check\_out\_date, total\_price, check\_in\_date, address, number\_of\_adults, where\_to
    \\Media: title, directed\_by, subtitles, genre
    \\Movies: theater\_name, movie\_name, price, show\_date, location, show\_time, number\_of\_tickets, genre, show\_type, street\_address
    \\Music: song\_name, year, album, artist, genre, playback\_device
    \\RentalCars: dropoff\_date, pickup\_time, pickup\_city, pickup\_date, total\_price, car\_type, car\_name, pickup\_location
    \\Restaurants: price\_range, restaurant\_name, city, has\_live\_music, serves\_alcohol, time, date, phone\_number, cuisine, street\_address, party\_size
    \\Salons: is\_unisex, average\_rating, city, appointment\_date, appointment\_time, stylist\_name, phone\_number, street\_address
    \\Dentists: dentist\_name, phone\_number, offers\_cosmetic\_services, city, appointment\_date, appointment\_time, address
    \\Doctors: doctor\_name, city, average\_rating, appointment\_date, appointment\_time, type, phone\_number, street\_address
    \\Travel: good\_for\_kids, category, attraction\_name, location, phone\_number, free\_entry
    \\Weather: city, temperature, date, precipitation, humidity, wind

    \hfill \\
    Further, following slots have categorical values:
    \\recipient\_account\_type: checking, savings
    \\fare\_type: Economy, Economy extra, Flexible
    \\(Travel) category: Place of Worship, Theme Park, Museum, Historical Landmark, Park, Tourist Attraction, Sports Venue, Shopping Area, Performing Arts Venue, Nature Preserve
    \\event\_type: Music, Sports
    \\seating\_class: Economy, Premium Economy, Business, First Class
    \\refundable: True, False
    \\airlines: United Airlines, American Airlines, Delta Airlines, Southwest Airlines, Alaska Airlines, British Airways, Air Canada, Air France
    \\show\_type: regular, 3d, imax
    \\playback\_device: TV, kitchen speaker, bedroom speaker
    \\(Doctors) type: Gynecologist, ENT Specialist, Ophthalmologist, General Practitioner, Dermatologist
    \\car\_type: Compact, Standard, Full-size
    \\price\_range: inexpensive, moderate, expensive, very expensive

    \hfill \\
    Further, following slots are boolean:
    \\has\_wifi, pets\_allowed, subtitles, offers\_cosmetic\_services, has\_laundry\_service, is\_unisex, good\_for\_kids, has\_live\_music, pets\_welcome, serves\_alcohol, is\_redeye, furnished, free\_entry
    
\end{tcolorbox}
\vspace{-1em}
\begin{tcolorbox}[colback=blue!15,colframe=blue!15,boxsep=-1mm,breakable]
    \scriptsize
    ---\\
    \hfill \\
    \texttt{{\color{blue}\{\% if} model\_name == {\color{red}{\dbq}llama\dbq\xspace}{\color{blue}\%\}}}\\
    Follow the following format.\\
    \texttt{\color{blue}\{\% else \%\}}\\
    The examples are formatted as follows.\\
    \texttt{\color{blue}\{\% endif \%\}}\\
    
    Dialogue:\\
    <user\_utterance>\\
    
    Applicable Standing Instructions:\\
    <applicable\_standing\_instructions>\\ 

    Tagged Standing Instructions:\\
    Tagged standing instructions\\
    
    API Calls: \\
    API calls to solve the user task\\

    ---\\

    \texttt{{\color{blue}\{\% if} model\_name == {\color{red}{\dbq}llama\dbq\xspace}{\color{blue}\%\}}}\\
    You are given several independent examples of the task:\\
    \texttt{\color{blue}\{\% endif \%\}}
\end{tcolorbox}

\noindent Example template: 
\begin{tcolorbox}[colback=red!10,colframe=red!10,boxsep=-1mm,breakable]
    \scriptsize
    \texttt{{\color{blue}\{\% if} split == {\color{red}{\dbq}test\dbq\xspace} {\color{blue}and} model\_name == {\color{red}{\dbq}llama\dbq\xspace}{\color{blue}\%\}}}\\
    Given the examples above, output only the API calls for the following example with no additional text:\\
    \texttt{\color{blue}\{\% endif \%\}}\\
    
    Dialogue:
    \\ \texttt{{\color{blue}\{\{} user\_utterance {\color{blue}\}\}}}\\
    \hfill \\
    Applicable Standing Instructions:
    \\ \texttt{{\color{blue}\{\{} applicable\_instructions | join({\color{red}{\dbq}$\backslash$n> {\dbq}}) {\color{blue}\}\}}} \\
    \hfill \\
    Tagged Applicable Standing Instructions:
\end{tcolorbox}

\noindent Target template:

\begin{tcolorbox}[colback=red!10,colframe=red!10,boxsep=-1mm,breakable]
    \scriptsize
    \texttt{{\color{blue}\{\{} tagged\_applicable\_instructions | join({\color{red}{\dbq}$\backslash$n> {\dbq}}) {\color{blue}\}\}}} \\
    \hfill \\
    API Calls: \\
    \texttt{{\color{blue}\{\{} target\_api\_calls | join({\color{red}{\dbq}$\backslash$n{\dbq}}) {\color{blue}\}\}}}
\end{tcolorbox}

\subsection{Tagger Prompt} \label{app:data-aug-tagger}

\noindent System prompt template:

\begin{tcolorbox}[colback=gray!20,colframe=gray!20,boxsep=-1mm,breakable]
    \scriptsize
    Create a sentence tagging model capable of identifying and labeling actions and their associated details within sentences. Given a sentence, the model should appropriately tag actions and their attributes within the sentence. \\
    The output should include all of the tokens from the original sentence, as well as action tags such as \text{[IN:ACTION ]} and nested tags denoting specific attributes, like \text{[SL:ATTRIBUTE value]}. \\
    Ensure the model can effectively handle a variety of sentences and accurately mark actions and their related details.\\
    \hfill\\
    \textnormal{Every action name has the format of "GET\_X", where X denotes the domain name.}\\
    Every action has a list of associated attributes. Only those attributes can be present inside the action tag.
    
\end{tcolorbox}
\vspace{-1em}
\begin{tcolorbox}[%
        enhanced, 
        colback=yellow!30,colframe=yellow!30,boxsep=-1mm,breakable]
    \scriptsize
    The list of the available function names is presented below, followed by possible slot names.
    \\Some of the possible API calls include functions:
    \\GetBanks: handling all the banking information (recipient\_account\_name, amount, recipient\_account\_type)
    \\ GetBuses: finding and booking bus tickets and routes (origin, departure\_date, fare\_type, transfers, price, group\_size, destination, departure\_time)
    \\ GetEvents: finding and booking events (event\_name, city, category, number\_of\_tickets, time, date, venue\_address, event\_type)
    \\ GetFlights: finding and booking flights (origin, inbound\_arrival\_time, is\_redeye, outbound\_departure\_time, outbound\_arrival\_time, inbound\_departure\_time, return\_date, airlines, seating\_class, refundable, number\_stops, departure\_date, fare, destination, passengers)
    \\ GetHomes: looking for property (pets\_allowed, visit\_date, address, property\_name, rent, number\_of\_baths, area, number\_of\_beds, furnished, phone\_number)
    \\ GetHotels: booking hotels (has\_wifi, average\_rating, check\_out\_date, price, pets\_welcome, number\_of\_days, location, check\_in\_date, phone\_number, number\_of\_rooms, street\_address, hotel\_name)
    \\ GetHouseStays: booking temporary accommodation (rating, phone\_number, has\_laundry\_service, check\_out\_date, total\_price, check\_in\_date, address, number\_of\_adults, where\_to)
    \\ GetMedia: searching for online media (title, directed\_by, subtitles, genre)
    \\ GetMovies: searching for cinema tickets (theater\_name, movie\_name, price, show\_date, location, show\_time, number\_of\_tickets, genre, show\_type, street\_address)
    \\ GetMusic: finding songs (song\_name, year, album, artist, genre, playback\_device)
    \\ GetRentalCars: booking rental cars (dropoff\_date, pickup\_time, pickup\_city, pickup\_date, total\_price, car\_type, car\_name, pickup\_location)
    \\GetRestaurants: finding and booking restaurants (price\_range, restaurant\_name, city, has\_live\_music, serves\_alcohol, time, date, phone\_number, cuisine, street\_address, party\_size)
    \\GetSalons: finding hair salons (is\_unisex, average\_rating, city, appointment\_date, appointment\_time, stylist\_name, phone\_number, street\_address)
    \\GetDentists: finding dentists (dentist\_name, phone\_number, offers\_cosmetic\_services, city, appointment\_date, appointment\_time, address)
    \\GetDoctors: finding doctors (doctor\_name, city, average\_rating, appointment\_date, appointment\_time, type, phone\_number, street\_address)
    \\GetTravel: finding attractions (good\_for\_kids, category, attraction\_name, location, phone\_number, free\_entry)
    \\GetWeather: getting weather information (city, temperature, date, precipitation, humidity, wind)

\end{tcolorbox}
\vspace{-1em}
\begin{tcolorbox}[colback=gray!20,colframe=gray!20,boxsep=-1mm,breakable]
    \scriptsize
    Check that the output fits all of the criteria above, and all of the tags are correctly placed (for example, \text{[SL: ]} tags must be inside the \text{[IN: ]} tags)\\
    Pay special attention to the attribute names and function names, check that none of the attribute names are mixed up (for example, some functions have similar attributes: city/location, make sure you are using the correct name)\\
    Check that all of the tokens from the original untagged sentence are present and are in the correct order. \\
    Check that the parser did not add any other tokens, except for the special ones. \\
    Make sure that the attribute values inlcude only the necessary information (for example, `\text{[SL:EVENT\_TYPE event type is Music]}' is incorrect and should be `event type is \text{[SL:EVENT\_TYPE Music]}').
\end{tcolorbox}

\noindent Example template: 
\begin{tcolorbox}[colback=red!10,colframe=red!10]
    \scriptsize
    \texttt{{\color{blue}\{\{} instruction {\color{blue}\}\}}}
\end{tcolorbox}

\noindent Target template: 
\begin{tcolorbox}[colback=red!10,colframe=red!10]
    \scriptsize
    \texttt{{\color{blue}\{\{} tagged\_instruction {\color{blue}\}\}}}

\end{tcolorbox}

\end{document}